\RequirePackage[l2tabu, orthodox]{nag}

\documentclass[12pt,phd,a4paper,oneside]{ucl_thesis}

\usepackage{blindtext}

\usepackage{emptypage}

\usepackage{graphicx}

\usepackage{float}

\usepackage{amsmath}

\usepackage{gensymb}
\usepackage{textcomp}

\usepackage{setspace}
\usepackage{multirow}

\usepackage{bibentry} 

\usepackage[format=hang,font=small,labelfont=bf]{caption}

\usepackage{etoolbox}

\usepackage{amsfonts}
\usepackage{cancel}
\usepackage[ruled]{algorithm2e}

\DeclareMathAlphabet{\mathcal}{OMS}{cmsy}{m}{n}

\usepackage{bibentry}
\makeatletter\let\saved@bibitem\@bibitem\makeatother
\usepackage[pdftex,hidelinks]{hyperref}
\makeatletter\let\@bibitem\saved@bibitem\makeatother
\makeatletter
\AtBeginDocument{
    \hypersetup{
        pdfsubject={Thesis Subject},
        pdfkeywords={Thesis Keywords},
        pdfauthor={Author},
        pdftitle={Title},
    }
}
\makeatother

\begin{document}
%
%
%
%
%
\makeatletter
\renewcommand {\@degree@string}{Master of Research}
\renewcommand {\@supervisor@string}{Dr Ben Calderhead}
\makeatother
\title{Efficient Exploration in Deep Reinforcement Learning: A Novel Bayesian Actor-Critic Algorithm}
\author{Nikolai Rozanov}
\department{Department of Computer Science}
\maketitle
\makedeclaration

\begin{abstract} 
Reinforcement learning (RL) and Deep Reinforcement Learning (DRL), in particular, have the potential to disrupt and are already changing the way we interact with the world. One of the key indicators of their applicability is their ability to scale and work in real-world scenarios, that is in large-scale problems. This scale can be achieved via a combination of factors, the algorithm's ability to make use of large amounts of data and computational resources and the efficient exploration of the environment for viable solutions (i.e. policies). 

In this work, we investigate and motivate some theoretical foundations for deep reinforcement learning. We start with exact dynamic programming and work our way up to stochastic approximations and stochastic approximations for a model-free scenario, which forms the theoretical basis of modern reinforcement learning. We present an overview of this highly varied and rapidly changing field from the perspective of Approximate Dynamic Programming. We then focus our study on the short-comings with respect to exploration of the corner-stone approaches (i.e. DQN, DDQN, A2C) in deep reinforcement learning. On the theory side, our main contribution is the proposal of a novel Bayesian actor-critic algorithm. On the empirical side, we evaluate Bayesian exploration as well as actor-critic algorithms on standard benchmarks as well as state-of-the-art evaluation suites and show the benefits of both of these approaches over current state-of-the-art deep RL methods. We release all the implementations and provide a full python library that is easy to install and hopefully will serve the reinforcement learning community in a meaningful way, and provide a strong foundation for future work.
\end{abstract}

\begin{acknowledgements}
First and foremost, I want to thank my mother. (Thanks mom). For your love and dedication. I also want to thank my father and my siblings. I would not have been where I am without you. Your support and incredible inspiration have brought me here.\\
Dr Ben Calderhead, apart from the critical thoughts, advice, guidance and trust I have received from you in the capacity of my supervisor, it is your human character that I value the most in this time that I had the pleasure of working alongside you. Thank you. \\
I also shall not forget all the friends with whom I have shared a common path so far. Particularly, Maro\v{s} Jan\v{c}o, Jean-Fran\c{c}ois Ton, Petru Constantinescu, Micha\l{} 
Maliczowski, Dorothee Dober, Eliana Fausti and Flavian Manea - thank you for the countless hours of love and happiness. \\

\noindent -- To working for the better of mankind.\\

\noindent Finally, I would like to acknowledge the studentship awarded by UKRI with the award reference number 1923151 that facilitated this research. 
\end{acknowledgements}

\setcounter{tocdepth}{2} 

\tableofcontents

\chapter{Introduction}
\label{chapterlabel1}
Technology and mankind have evolved together in an endless cycle. This work aspires to be a tiny step in the same direction; it should contribute to making applied systems possible that can be used in the real world. The nature of the methods, results and findings described here is for the purpose of building systems that can adapt to dynamic and uncertain environments and act optimally in these. This class of methods is referred to as sequential decision making nowadays more commonly known as reinforcement learning. There is a rich literature on such problems and methods stemming from different fields and under different names: Control Theory \cite{bellman1966dynamic},  Reinforcement Learning \cite{Sutton2018}, Neuro-dynamic Programming \cite{bertsekas1996neuro}, Operations Research \cite{ortega2004control}, Neuroscience \cite{mnih2015human}, Game Theory and Economics \cite{seierstad1986optimal} and many more. The full range of applications of sequential decision-making methods is probably innumerable, as most of the human activity and many natural occurrences are precisely of the nature of sequential decision making. Some examples are medical trials, where the dosage, frequency and whether to continue needs to be decided on a regular basis given the prognosis of a patient, or management of an energy supply chain, where each power plant's output needs to be constantly regulated and directed, or more modern applications such as building autonomous vehicles or robotics control more generally, as well as optimal routing and warehouse management, which are widely needed in industry and logistics, among others. Naturally, many more scientific discoveries and contributions need to happen in order for us to be able to employ artificially intelligent systems across all of these without human supervision - if such a thing will ever be a reality.

The current progress in the field of sequential decision making comes from the multi-disciplinary field of machine learning and in particular from reinforcement learning (RL). The unprecedented availability of large-scale compute and data triggered a wave of exploration into methods that scale with both of these parameters. It turned out that deep learning (DL) \cite{lecun2015deep} was precisely such a method that scaled very well with available compute and data, as it doesn't require manual feature engineering but learns to discover salient features automatically. Deep learning was the necessary catalyst for another wave of Artificial Intelligence (AI) research. The actual dawn can probably be marked by \cite{krizhevsky2012imagenet}. This metaphorical explosion of methods, applications and advances in machine learning also brought its fruits to sequential decision making in the form of Deep RL. The most notable results are those where human performance, assumed to be the gold standard, was overtaken by these algorithms. In particular, examples include algorithms outperforming humans on many of the Atari2600 Games \cite{mnih2015human}, winning against one of the top-ranked Go players \cite{silver2016mastering}, winning against professionals in an \textbf{\href{https://openai.com/blog/openai-five/}{e-sport}}, which used Deep RL \cite{schulman2017proximal}, etc.. However, as is noted by many of the authors themselves, these accomplishments required enormous amounts of computational power and the ability to simulate the given environments. This naturally poses a significant obstacle to applying these advances in real-world scenarios, where the environment cannot be simulated millions or billions of times. Hence, at least at this stage in the evolution and the distribution of this technology, the main obstacle lies in the data efficiency of these methods. This translates to the very central question of exploration vs. exploitation in sequential decision-making problems and the motivation behind our work. 

\section{Thesis Outline}
This work can be seen as a combination of a survey of recent advances, methodological proposals, empirical evaluations of the core ideas and building blocks for follow-on research. Deep reinforcement learning is a highly varied and rapidly evolving field. In this work, apart from our other contributions, we provide a distinct overview of the most recent methods and advances. The perspective of this overview takes a more foundational view from the perspective of exact dynamic programming and its approximations. We elaborate on the key results and stochastic approximations to the exact methods developed by Bellman et. al., which serves as one of the two foundations that motivate the choice and development of our own algorithm. The second foundation lies in the topic of exploration vs. exploitation within the RL and Deep RL setting. The majority of state-of-the-art methods conduct exploration and learning via the  $\epsilon$-greedy approach or the Boltzmann distribution over the decision or actions space, however, it has been shown that these strategies can lead to worst-case performance \cite{osband2017deep}. Our work builds on previous work \cite{osband2017deep}, which studies the shortcoming of these methods and proposes Bayesian methods based on Thompson sampling \cite{thompson1933} to improve the data efficiency. Our theoretical contribution lies in proposing a novel Bayesian actor-critic algorithm that combines the theoretical foundation of actor-critic methods as well as the strength of Bayesian exploration using Thompson sampling. Finally, we implement and study these individual benefits empirically and compare them to current state-of-the-art models.

The outline of our work is as follows. 
\hyperref[chapterlabel2]{ \textbf{ Chapter 2 - Background and Motivation } } introduces the state-of-the-art of the field and presents the main concepts: Markov decision processes (MDP), Value Iteration, Temporal Difference learning, stochastic approximation and Bayesian Deep Learning. 
\hyperref[chapterlabel4]{ \textbf{ Chapter 3 - Design, Methods and Experiment } } outlines our novel algorithm, the main experiments and describes how empirical validation was conducted. \hyperref[chapterlabel5]{ \textbf{ Chapter 4 - Analysis } } focuses on the results, where we empirically show that methods that guide exploration with tracked uncertainty estimates show promising results and that actor-critic algorithms show promising results in their convergence speed as well. \hyperref[chapterlabel6]{ \textbf{ Chapter 5 - Conclusion \& Outlook } } discusses how the results fit into the bigger picture and describe how this work serves as a foundation for future work.



\chapter{Background and Motivation}
\label{chapterlabel2}

This chapter gives a full overview of the methods used in contemporary RL. We study the theoretical foundation that underlies these methods, as well as the clever stochastic approximation methods that allow these methods to scale to modern-day requirements. 
We particularly emphasise the stochastic approximation techniques as these are quite often omitted in today's analysis and discussion within the community. We present the Q-learning algorithm and briefly discuss the implications that non-linear function approximators, such as neural networks, have on the convergence properties of this algorithm. We also discuss a very different set of algorithms, which are called policy gradient methods. These set the foundation for our actor-critic method.
\section{Recent Advances in Reinforcement Learning}

Before diving into the theoretical foundations of RL, it is beneficial to outline some of the most striking advances in RL and their unbelievable results. This should give further motivation for the importance of reinforcement learning, as well as set clear targets when investigating the theory as to what needs to be understood. 

\noindent The most significant advances in (deep) RL are very likely: Alpha Go Zero \cite{alphaGoZero}, which beat one of world's best Go players, AlphaZero \cite{alphaZero}, which learned Go, Chess and Shogi among others only through self-play, OpenAI's Dota Five using the PPO algorithm \cite{schulman2017proximal}, DQN applied to Atari games \cite{mnih2015human}, Superhuman scores on Montezuma's Revenge without imitation learning \cite{burda2018exploration}, and the application of a pure simulation trained RL algorithm in a real-world robot \cite{andrychowicz2018learning}. The following should serve as a quick summary for future reference:

\noindent \textbf{DQN}\\
The main accomplishment, as stated in the paper as well, is the possibility of using highly nonlinear function approximators to approximate the Q-function. The method that allows this is the use of Experience Replay, which allows the network to learn outside of the current behaviour; the use of two networks, which allows again further decoupling from learning and acting; and finally the Least Square Temporal Difference Error:\footnote{The Temporal Difference Error, as well as the notation used, will be elucidated in later parts of the work.}
\begin{align*}
    \mathcal{L}(\theta_i) = \mathbb{E}_{(x,R) \sim \text{Experience Replay}} \bigg[ \bigg(R + \max_{a\in\mathcal{A}}\gamma Q(x,a,\theta_i') - Q(x,a,\theta_i)\bigg)^2 \bigg]
\end{align*}
where $\theta$ represents the parameters of the first Neural Network and $\theta'$ represents the second, decoupled Neural Network. $\theta'$ is synchronised with $\theta$ every C time-steps. 

\noindent \textbf{Alpha Go Zero}\\
The Alpha Go work, which accumulated in three papers accomplished many unprecedented things. In particular, it managed to outperform a professional human player in Go for the first time in human history.\\ 
The algorithmic details of Alpha Go go beyond the scope of this work, as they utilise additional features such as Monte Carlo Tree Search to improve upon standard learning in Reinforcement Learning settings. 

\noindent \textbf{PPO applied to Dota 2}\\
Dota 2 is a complicated multi-player online game. Being good at it involves team-work, short term control as well as long term planning. The fact that a single algorithm manages to accomplish this without human demonstration is a clear demonstration that RL algorithms are already able to solve problems that resemble the real world. While the full set-up for Open AI's team is out of scope for our work, the PPO algorithm \cite{schulman2017proximal} is briefly discussed.

\noindent The proximal policy optimisation algorithm is based on learning a single network that has both the task of predicting the value estimates for states as well as acting as the policy. This builds on some of the ideas in Alpha Go. The empirical results suggest that such joining of purposes into a single network has an advantageous effect. This method falls into the category of actor-critic methods. 

\noindent \textbf{Exploring Montezuma's Revenge}\\
Montezuma's revenge is a very challenging game in so far that rewards are very sparse. I.e. one has to take many actions in order to see any change in the score of the game. Naturally, $\epsilon$-greedy exploration methods are predestined to not accomplish anything in such scenarios, as the chances of randomly winning a game are diminishingly low over larger time horizons. The clever idea of Open AI's team was to build an internal reward signal for the agent that rewards curiosity, which they accomplish by trying to predict the next state, i.e. by trying to learn the transition kernel $\mathcal{P}$ and always trying those actions of the game where the predictions were the worst. Using this method Open AI's agent manages to be the first algorithm trained entirely through self-play to outperform humans in Montezuma's Revenge.

\noindent \textbf{Learning in Simulation applying in the World}\\
The final work that will be briefly discussed is again Open AI's contribution to the field. They managed to take an agent trained in a simulated environment and apply it in the real world. This is fascinating from many perspectives; firstly, this brings reinforcement learning agents so much closer to the real world, secondly, this demonstrates a method of how one can leverage large scale computation to solve problems in the real world. The main technical idea lies in training the agent with many variations of a similar environment. What this means is that the physical properties of the robot such as friction, momentum etc. were left as random variables and the agent learned to solve all of the versions of the environment, hence when it was applied in the real world it was merely solving one of the variations it learned to solve.

This variety of algorithms leaves us with a natural question of which ones to use, when and why? To understand and decide between this variety of algorithms it is crucial to understand their theoretical motivation. This theoretical foundation is even more important when one wants to design and contribute to new algorithms. Furthermore, all of these advances have a single thing in common and this is the incredible amount of computation that had to be used to train these systems. This is only applicable for a handful of problems, however, building systems that can interact in everyday scenarios with limited computation available is still a significant challenge and therefore work on computationally more efficient methods is one of the key directions for the development of such systems. 

\section{Preliminaries}
Before diving into an in-depth discussion on the various theoretical topics we want to establish consistent notation and outline important underlying assumptions of the problem at hand. 

In a rather general sense, a single agent\footnote{Multi-Agent Sequential Decision making is another topic altogether and brings its own challenges such as asymmetric information flow, see for example Witsenhausen's counterexample.\cite{Witsenhausen1968ACI} } sequential decision-making problem can be described as a continuous control problem with an underlying stochastic process governing the environment dynamics, with some functional of the continuous stochastic reward (or cost) function\footnote{Without loss of generality control problems can be stated either as reward maximising or cost minimising problems, with reinforcement learning conventionally using reward functions.} being the objective function. I.e.:
\begin{align*}
   \dot{x}(t) &= P(x(t), a(t), t) \\
    V &= \Phi(x(t_0),x(T),t_0, T) + \int_{t=t_0}^{t=T} \mathcal{R} \bigg(x(t), a(t), t\bigg) dt \\
    o &= \Pi x
\end{align*}
where x is the state of the environment $ \forall t\in [t_0, T]$. $P$ is a time indexed stochastic process, $ \Phi $ is some reward function at the boundary conditions acting as an offset. $\mathcal{R}$ as the main reward function or distribution, $V$ acts as the objective function that needs to be optimised by the agent's control signal $a$ and $o$ is the observation signal that the agent receives and can use to optimise its actions. Now, in our case we will be studying this problem with several assumptions. 
\begin{enumerate}
\item \textbf{Discrete Problem} - actions (or control) as will be discrete, i.e.: $a \in \{0,1,2,...,n\}$ and time t will be discrete (, i.e. $t\in\{t_0,...,t_T\}$. Time t is not necessarily assumed to be finite.
\item \textbf{Fully Observable} - the agent observes x directly, i.e.: $\Pi$ is assumed to be the identity function.
\item \textbf{Strict Stationarity} - we will assume that the stochastic transition function $P$ is time independent.
\end{enumerate}

\begin{figure}[h]
    \centering
    \includegraphics[width=\textwidth]{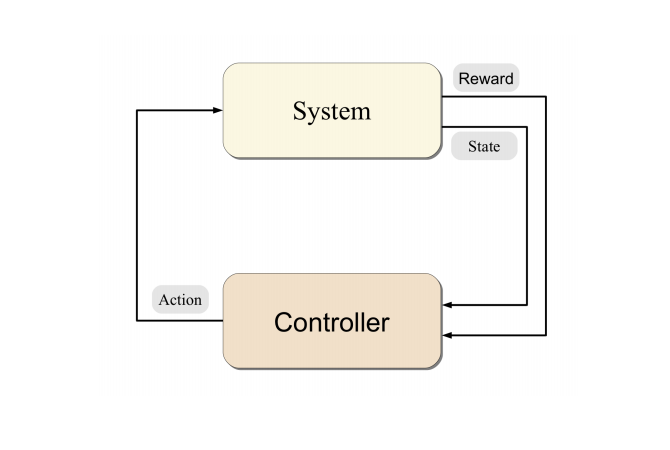}
    \caption{The control setting: Agent interacting with an environment and receiving a reward signal. }
    \label{fig:state_reward}
\end{figure}

These assumptions are standard in the reinforcement learning community \cite{osband2017deep,szepesvari2010algorithms}. However, even apart from this, these assumptions allow for an incredible variety of problems and applications in the real world. Perhaps the two most significant assumptions in this work, as well as in the majority of recent deep RL research, are the stationarity assumption and the full observability. The resulting problem with these assumptions can be expressed as a Markov Decision Process (MDP).

\section{Markov Decision Process (MDP)}
We start off by defining a (finite) Markov Decision Process (MDP). An MDP is a tuple $ \mathcal{M} = (\mathcal{X},\mathcal{A},\mathcal{P},\mathcal{R} \rho_0) $. $ \mathcal{X} $ is a finite state space and without loss of generality can be assumed to be $ \mathcal{X}=\{0,1,2,...,S\}$, with state $x=0$ being the terminal state\footnote{Note that in some literature the terminal state is treated separately, including the system dynamics. These methods are equivalent and in our case x=0 can and should be treated as an absorbing state that is connected to every other.} in case of finite-horizon MDPs, $ \mathcal{A} $ is a finite actions space and $\mathcal{R} \subset \mathbb{R}$ is the domain of the reward. $\mathcal{P} $ is the transition probability kernel that governs the dynamics and assigns a probability to every state and reward $x\in \mathcal{X} \times \mathcal{R}$ for every state and action $(x,a) \in \mathcal{X} \times \mathcal{A}$, as $(x',R) \sim \mathcal{P}( \cdot |x,a )$. Finally, $\rho_0$ defines the distribution of the initial state $x(0)=x_0 \sim \rho_0$.\\
Given the above definition of the MDP, an agent interacts with it sequentially by taking a decision at every time step $t$.  The following recursive equations summarise this:
\begin{align}
    x_0 &\sim \rho_0 \\
    (x_{t+1}, R_{t+1}) &\sim \mathcal{P}(x_{t+1}, R_{t+1} | x_t, a_t) \\
\end{align}
An episode is defined to be a chain of $(x_0,a_0, x_1, R_1, a_1, ..., x_T, R_T)$ where $x_T=0$ is the terminating state. In case of infinite-horizon MDPs there is no terminal state.\\
\noindent The finite-horizon return is defined as:
\begin{align}
    G = \sum_{t=1}^{T} R_t 
\end{align}
Or correspondingly in the infinite-horizon case:
\begin{align}
     G = \sum_{t=1}^{\infty} \gamma^t R_t  \quad \text{for } \gamma \in (0,1) 
\end{align}
The second definition incorporates the first with $R_t=0$ for $t>T$ and $\gamma=1$, $\forall t$. Hence, we will be using the second to refer to both cases.
We will also define related terms and functions that are used when developing results and methods to solve MDPs. 

\noindent \textbf{State Transition Probabilities}\\
It is sometimes useful to just consider the state-transitions without the associated reward. The state-transition probabilities $p: \mathcal{X} \times \mathcal{X} \times \mathcal{A} \rightarrow [0,1]\subset \mathbb{R}$ are given by:
\begin{align*}
    x_{t+1} \sim p( x_{t+1} | x_t, a) = \int_{R\in \mathcal{R}} \mathcal{P}(x_{t+1}, R | x_t, a) \, dR
\end{align*}

\noindent \textbf{State-Action Expected Reward}\\
the state-action expected rewards $r: \mathcal{X} \times \mathcal{A} \rightarrow \mathcal{R} \subset \mathbb{R}$ are given by:
\begin{align*}
    r(x,a) = \mathbb{E} \bigg[ R | x,a\bigg]
\end{align*}
Where $R$ comes from $(x', R) \sim \mathcal{P}(x,R|x,a)$

\noindent \textbf{Policy}\\
A policy is a conditional distribution over the action space $a \sim \pi(\cdot| x)$, usually denoted $\pi(x)$. A deterministic policy is a mapping $\pi : x \rightarrow a $.

\noindent \textbf{Value Function}\\
The (state) value function given policy $\pi$, denoted $V^{\pi}$, is defined as a conditional expectation of the return (2.5), i.e.:
\begin{align}
    V^{\pi}(x) = \mathbb{E} \bigg[\sum_{t=1}^{\infty} \gamma^t R_t | x_0=x\bigg] \label{eq:value_equation} 
\end{align}
Here the expectation is taken over sequences of observations by following the modified transition kernel $\mathcal{P}^{\pi}(\cdot | x) = \sum_{a\in \mathcal{A}} \mathcal{P}(\cdot | x, a)\pi(a|x)$.

\noindent \textbf{State-Value Function (Q-Function)}\\
The state-value function given a policy $\pi$ is commonly known and denoted as $Q^{\pi}(x,a)$. It is a natural extension of the value function:
\begin{align}
    Q^{\pi}(x,a) = \mathbb{E} \bigg[\sum_{t=1}^{\infty} \gamma^t R_t | x_0=x, a_0=a\bigg] \label{eq:q_equation} 
\end{align}
Here the expectation is taken in the same way as for the Value function.


\section{Exact Dynamic Programming}
In this section, we want to discuss what it means to perform well in MDPs. In particular, we discuss the Bellman Equations and Bellman optimality equation. Then we establish that given that one knows the true dynamics of the environment one can use Dynamic Programming to solve the MDP, we conclude this section by discussing the obvious short-comings of these methods. 

\noindent \textbf{Optimal Value functions}\\
The optimal value function, denoted $V^{*}(x)$, is one that has the highest possible value for every state x. It is defined by:
\begin{align}
    V^{*}(x) = \max_{\pi} V^{\pi}(x) \quad, \forall x
\end{align}
Similarly, for the state-action value function:
\begin{align}
    Q^{*}(x,a) = \max_{\pi} Q^{\pi}(x,a) \quad, \forall x \forall a
\end{align}
In this case taking the maximum over all possible policies does not pose a problem, since the optimal policy can be deterministic \cite{bellman1966dynamic} and hence there are finitely many possibilities. 

\noindent \textbf{Bellman Optimality Equation}\\
The Bellman Optimality Equation \cite{bellman1966dynamic} relates the value function to the state value function in the optimal case.  This relation also helps establish a recursive formulation of both the value function and state-action value function. The intuition behind the equation lies in the fact that the optimal policy also maximised the current step. I.e.:

\begin{align}
    V^{*}(x) &= \max_{a\in\mathcal{A}} Q^{*}(x,a) \label{eq:bellman_optimality_principle} \\
             &= \max_{a\in\mathcal{A}} \mathbb{E} \bigg[\sum_{t=1}^{\infty} \gamma^t R_t | x_0=x, a_0=a\bigg] \\
             &= \max_{a\in\mathcal{A}} \mathbb{E} \bigg[R_1 + \sum_{t=2}^{\infty} \gamma^t R_t | x_0=x, a_0=a\bigg] \\
             &= \max_{a\in\mathcal{A}} \mathbb{E} \bigg[R_1 + \gamma \sum_{t'=1}^{\infty} \gamma^{t'} R_{t'} | x_0=x, a_0=a\bigg] \\
             &= \max_{a\in\mathcal{A}} \mathbb{E} \bigg[ R_1 + \gamma  V^{*}(x_1)  | x_0=x, a_0=a \bigg] \\
             &= \max_{a\in\mathcal{A}} \bigg\{ r(x,a) + \gamma \mathbb{E}_{x_1 \sim p|(x,a)} \bigg[ V^{*}(x_{1})\bigg] \bigg\}\\
            &= \max_{a\in\mathcal{A}} \bigg\{  r(x,a) + \gamma \sum_{x_1\in\mathcal{X}}p(x_1|x,a) V^{*}(x_{1}) \bigg\} \label{eq:optim_value_equation} 
\end{align}

Similarly for $Q^{*}$ one can derive the Bellman optimality equation:
\begin{align}
     Q^{*}(x,a) &= r(x,a) + \gamma \sum_{x'\in\mathcal{X}} p(x'|x,a) \max_{a'\in\mathcal{A}} Q^{*}(x',a') \label{eq:optimal_q_equation}
\end{align}

\noindent \textbf{Optimal Policy}
Given the optimal state value function or state-action value function then the optimal policy is given by:
\begin{align}
    \pi^{*}(a | x) &= arg\,\max_{a\in\mathcal{A}} Q^{*}(x,a) \label{eq:optimal_policy_equation}
\end{align}

\subsection{Value Iteration}
We now want to introduce an algorithm that allows us to compute the optimal value function and therefore provide a method of computing the optimal policy.\\
Let us first write equation (\ref{eq:optim_value_equation}) more compactly in operator notation. We will refer to this operator as the Bellman optimality operator, i.e.:
\begin{align}
    (\ref{eq:optim_value_equation}) &= V^{*} = T^{*}V^{*} \label{eq:optim_value_operator}
\end{align}
This equation forms the basis of the value iteration algorithm. The core principle lies in defining an iterative scheme:
\begin{align}
    V_0(x) &= 0  & \forall x\in \mathcal{X} \\
    V_{n+1}(x) &= (T^{*}V_{n})(x) & \forall x\in \mathcal{X} 
\end{align}
The convergence of this scheme relies on the fact that $T$ is a contraction operator under the max norm. Hence, by Banach's fixed-point theorem, V converges to a unique point for every state x and therefore must coincide with the optimal value function for every state. We show in appendix \ref{appendixlabel1} that $T^{*}$ is indeed a contraction mapping in the finite dimensional case, i.e. that $||T^{*}(V) - T^{*}(V')||_{\infty} < ||V - V'||_{\infty}$, where the max norm is taken over the individual states.

Interestingly, these results both translate with minor technical conditions to more general settings, where the finite cardinality of the MDP does not hold anymore. Furthermore, it also generalises to asynchronous state updates, as opposed to updating all states at once. (cf. \cite{bertsekas1996neuro}, Chapter 2, Proposition 2.3. (Asynchronous Value Iteration)).\\
Finally, once the value iteration algorithm terminates we want to be able to recover an optimal policy. It is clear that we can use (\ref{eq:optimal_q_equation}) to express the optimal state-action value function in terms of the optimal value function:

\begin{align}
    Q^{*}(x,a) = r(x,a) + \gamma \sum_{x'\in \mathcal{X}} p(x'|x,a)V^{*}(x) \label{eq:state_value_in_terms_of_value}
\end{align} 
Using this result and the relationship between a state-action value function and a corresponding greedy deterministic policy (\ref{eq:optimal_policy_equation}) we obtain the optimal policy $\pi^{*}$.

\subsection{Policy Iteration}
Another way of looking at obtaining the optimal policy is by an iterative and alternating improvement of the policy and value function. For this, we need to first consider how given some policy $\pi$ one can evaluate this policy, i.e. how one can compute the corresponding value function (\ref{eq:value_equation}).

\noindent\textbf{Policy Evaluation Step} \\
It turns out that using a slight variation of the Bellman optimality operator (\ref{eq:optim_value_equation}) and using a very similar derivation we can obtain a recursive equation to calculate the value function for a given policy. Assuming a deterministic policy, we can derive it as follows:
\begin{align}
    V^{\pi}(x)  &= \mathbb{E} \bigg[\sum_{t=1}^{\infty} \gamma^t R_t | x_0=x\bigg] \\
                &= \mathbb{E} \bigg[\sum_{t=1}^{\infty} \gamma^t R_t | x_0=x, a_0=\pi(x) \bigg] \\
                &= \mathbb{E} \bigg[R_1 + \sum_{t=2}^{\infty} \gamma^t R_t | x_0=x, a_0=\pi(x)\bigg] \\
                &=  \mathbb{E} \bigg[R_1 + \gamma \sum_{t'=1}^{\infty} \gamma^{t'} R_{t'} | x_0=x, a_0=\pi(x)\bigg] \\
                &= \mathbb{E} \bigg[ R_1 + \gamma  V^{\pi}(x_{1})  | x_0=x, a_0=\pi(x) \bigg] \\
                &= r(x,\pi(x)) + \gamma \mathbb{E}_{x_1 \sim \mathcal{P}|(x,\pi(x))} \bigg[ V^{\pi}(x_{1})\bigg] \\
                &=  r(x,\pi(x)) + \gamma \sum_{x_1\in\mathcal{X}}\mathcal{P}(x_1|x,\pi(x)) V^{\pi}(x_{1}) \label{eq:value_operator_equation}
\end{align}
Again, writing this more compactly in operator form, which we will refer to as the Bellman operator, we get:
\begin{align}
    (\ref{eq:value_equation}) &= V^{\pi} = T^{\pi}V^{\pi} \label{eq:value_operator}
\end{align}
Similarly, we have the Bellman operator for the state-action function:
\begin{align}
     Q^{\pi}(x,a) &= r(x,a) + \gamma \sum_{x'\in\mathcal{X}} p(x'|x,a) Q^{\pi}(x',\pi(x')) \label{eq:q_operator_equation}
\end{align}
Hence, starting of with some policy $\pi_0$ and using the Bellman operator, we can use the following iterative algorithm to evaluate the given policy: 
\begin{align}
    V_0(x) &= 0 & \forall x\in \mathcal{X} \\
    V_{t+1}(x) &= T^{\pi}V_{t}(x) & \forall x \in \mathcal{X}
\end{align}
As in the optimal case, it turns out that this is also a contraction mapping with respect to the max-norm. A proof would be very similiar to the one given in Appendix \ref{appendixlabel1}. Therefore, again by the Banach's fixed-point theorem we have convergence guarantees.\\

\noindent\textbf{Policy Improvement Step}\\
Apart from the policy evaluation step, we also need to understand how we can improve the policy and furthermore how we can arrive at an optimal policy. 

\noindent Given a current policy $\pi$ and constructing a new policy with respect to the corresponding value function $V^{\pi}$ can be done by taking a greedy one-step look-ahead action, i.e.:
\begin{align}
    \pi'(a|x) = arg\max_{a\in\mathcal{A}} Q^{\pi}(x,a) \label{eq:policy_improvement_step}
\end{align}

\noindent It can be shown (e.g. \cite{Sutton2018}, Chapter 4, (4.7) ) that (\ref{eq:policy_improvement_step}) leads to an improved value function $V^{\pi'} \geq V^{\pi}$. Furthermore, if at some point the inequality becomes an equality $V^{\pi'}(x)=V^{\pi}(x)$ $\forall x \in \mathcal{X}$ then $\pi$ is the optimal policy. This can be easily shown. Starting with the fact that $\pi'$ optimises $Q^{\pi}$:
\begin{align}
    V^{\pi'}(x) &= \max_{a \in \mathcal{A}} Q^{\pi}(x,a)\\
    & = \max_{a \in \mathcal{A}}\bigg \{r(x,a) + \gamma \sum_{x'\in \mathcal{X}} p(x'|x,a)V^{\pi}(x) \bigg \} \\
    & = \max_{a \in \mathcal{A}} \bigg \{r(x,a) + \gamma \sum_{x'\in \mathcal{X}} p(x'|x,a)V^{\pi'}(x) \bigg \} \label{intermediate:policy_improvement} \\
    &= T^{*} V^{\pi'}
\end{align}
Where (\ref{intermediate:policy_improvement}) follows from the assumption that the policy improvement step did not lead to any change. $T^{*}$, as always, represents the Bellman optimality operator and therefore we see that $V^{\pi'}$ is a fixed point of the Bellman optimality operator, however, because it is a contraction it is a unique fixed point and thus $V^{\pi'}=V^{*}$.\\
\null \hfill $\square$

\subsection{Generalised Policy Iteration}
As a final method for exact dynamic programming, we will briefly outline a generalised form of both the value iteration and policy iteration algorithm. One can see from the general format of the policy iteration algorithm that the policy improvement step needs a converged and correct value function for a particular policy, hence between any two policy improvement steps policy iteration requires the computation of the value function. This incurs a relatively large cost. On the other hand, value iteration requires one full convergence of the value function to the optimal value function. Furthermore, if one looks carefully then one sees that value iteration incorporates a policy improvement step at every stage. It turns out that one can do policy improvement steps before a full convergence of the value function $V^{\pi}$ and at any step in between. The proof is given in \cite{bertsekas1996neuro}, Chapter 2, Proposition 2.5. (Asynchronous Policy Iteration). Hence, the fully general form of value and policy iteration is given by:\\

\begin{algorithm}[H]
\setstretch{1}
\SetKwInOut{Input}{Input}\SetKwInOut{Output}{Output}
\Input{MDP($\mathcal{X},\mathcal{A},\mathcal{R},\mathcal{P},\rho_0$), T, $\epsilon$, N}
\Output{$\pi^{*}_{\epsilon}$\footnote{We denote by $X_{\epsilon}$ any function or value that is within $\epsilon$ of the optimal solution}, $V^{*}_{\epsilon}$ }
\BlankLine
//initialisation \;
 $V(x)\leftarrow 0$,   $\forall x\in\mathcal{X}$ \; 
 $\pi(x)\leftarrow rand$\footnote{Here $rand(a,n)$ returns an element of type a of cardinality n or 1 if n is not specified.}$(a)$, $a\in\mathcal{A}, \forall x \in\mathcal{X}$ \;
 \BlankLine
    \While{$||V-T^{*}(V)||_{\infty}>\epsilon$}{
        //Sample random states to update value and policy \;
        $\mathcal{X}' \leftarrow rand(X,N)$  $X\subset \mathcal{X}$\;
        $\mathcal{X}'' \leftarrow rand(X,N)$  $X\subset \mathcal{X}$\;
        \BlankLine
        //Value Update\;
        \For{$i\leftarrow 0$ \KwTo $T-1$}{
            \For{$x\in \mathcal{X}'$}{
                $V(x) \leftarrow T^{\pi}V(x)$ \;
            }
        }
        \BlankLine
        // Policy Update\;
        \For{$x\in \mathcal{X}''$}{
            $\pi(x) \leftarrow arg\max_{a \in\mathcal{A}} Q(x,a)$ \;
        }  
    }
 \caption{Generalised Policy Iteration}
 \label{alg:generalise_policy_iteration}
\end{algorithm}
\null 
 \noindent This is indeed a generalised policy iteration algorithm, as setting $T=1$ gives us value iteration, while setting $T=\infty$ gives us policy iteration.

\subsection{Summary and Shortcomings}
In this section, we have seen ways and methods to "solve'' a finite Markov Decision Process. Overall the method relies on evaluating policies with the help of value functions and then updating the policy given a greedy policy improvement step. These results extend beyond the finite case of MDPs \cite{bertsekas1996neuro,Sutton2018}. However, in practice these exact methods have two major shortcomings:
\begin{enumerate}
    \item Requirement of full knowledge of the MDP
    \item Tractability issues in large state or action spaces 
\end{enumerate}

\noindent\textbf{Full Knowledge of the MDP}\\
In reality, it is very rarely the case that the true dynamics of the system are known. In fact, for most real-world problems either the dynamics are not known or are incredibly hard to formulate correctly. Prominent examples are self-driving cars, robotics, dialogue systems, finance etc. In such cases, it would be impossible to apply these methods directly. In traditional control theory and in general when one wants to use these exact methods one needs to estimate the dynamics first \cite{deisenroth2011pilco,kaiser2019modelbased} and then use DP to solve the problem.

\noindent\textbf{Tractability in large state or action spaces}\\
General policy iteration requires that every state is iterated many times. Given for example even the game of Backgammon this is $10^{20}$, which clearly is not tractable. For the game of Go, this is a significantly larger number\cite{silver2016mastering}. Real-world scenarios such as self-driving cars, dialogue systems (such as Siri or Alexa), etc. have even larger state-spaces and even possibly countably many.

\noindent\textbf{Conclusion}\\ 
What should be noted is that Dynamic Programming (DP) as a method is not in-efficient, rather the scalability issues come from the size of the problems. In fact, it can be shown that DP finds the optimal solution in polynomial time in the actions and state-space sizes (i.e. $|\mathcal{X}|$ and $|\mathcal{A}|$)  \cite{Sutton2018}, Chapter 4, Section 4.7. Whereas the total policy space size is $|\mathcal{X}|^{|\mathcal{A}|}$. Hence, DP is exponentially faster than direct search algorithms would be. Nevertheless, we need methods that work well under large state and action spaces as well as methods that work without the knowledge of the true dynamics. We investigate this in the next section.

\section{Stochastic Approximations}
We have seen that exact dynamic programming, although with good theoretical guarantees and results, is still not applicable in real-world scenarios due to intractability and violation of certain assumptions, most importantly the complete knowledge assumption. In this section, we want to present and review several stochastic approximation methods that work well under imperfect knowledge and information conditions, as well as can be applied as approximate solvers when the full problem is not tractable. 

\noindent The general type of these stochastic approximation algorithms is of an iterative nature. In particular, we will be considering the following setting: 
\begin{align}
    Tx=x 
\end{align}
where T is some mapping from $\mathbb{R}^n$ into itself, and we want to find the fixed point. Furthermore, we will assume that we do not have access to the exact form of T, but rather we have access to a random variable s, via sampling or simulation:
\begin{align}
    s = Tx+w
\end{align}
In this case, w is some random noise term. The general form of estimators that we will be studying is:
\begin{align}
    x_{t+1} = (1-\alpha) x_t + \alpha (Tx_t+w) \label{eq:general_approximation_method}
\end{align}
To see that this allows for rather general and interesting problems, suppose we are interested in:
\begin{align}
    x = \mathbb{E} \bigg [f(x,z) \bigg ] \label{eq:x_equal_expectation}
\end{align}
Where the additional variable z is distributed as $z \sim p(z|x)$. We will further assume that f is known, but that the conditional distribution is not. The final assumption is that we have access to samples $\tilde{z_1},...,\tilde{z_k}$. This is a rather general setting and can also be easily applied in supervised learning settings by setting f to be the objective function and z to be the target. As k becomes large the sample mean:
\begin{align*}
    \frac{1}{k}\sum_{i=1}^{k}f(x,\tilde{z}_i)
\end{align*}
will converge to the desired mean and we could apply deterministic algorithms to solve for x, however, we will also explore alternatives, such as for example when k=1. 
\begin{align}
    x_{t+1} = (1-\alpha) x_t + \alpha f(x_t,\tilde{z}_t) \label{eq:robbins-monro}
\end{align}
In fact, this particular version of the algorithm is known as the \textit{Robbins-Monro} stochastic approximation algorithm. We can rewrite (\ref{eq:robbins-monro}) to recover our desired format (\ref{eq:general_approximation_method}).
\begin{align*}
    x_{t+1} = (1-\alpha) x_t + \alpha \bigg [ \mathbb{E}\Big[ f(x_t,z) \Big] + f(x_t,\tilde{z}) - \mathbb{E}\Big[ f(x_t,z) \Big] \bigg ]
\end{align*}
We can now set the operator T to be:
\begin{align*}
    Tx = \mathbb{E}\Big[ f(x,z) \Big]
\end{align*}
and the noise w to be:
\begin{align*}
    w = f(x,\tilde{z}) - \mathbb{E}\Big[ f(x,z) \Big]
\end{align*}
So far we have not discussed the step-size (or alternatively interpolation parameter) $\alpha$. In fact, $\alpha$ has to be iteration dependent as well, i.e. $\alpha_t$ and furthermore the conditions for these algorithms to converge under various technical assumptions are known as the Robbins-Monro (RM) conditions. 
\begin{align}
    \sum_{t=0}^{\infty} \alpha_t &= \infty  \label{def:rm_cond_1}\\
    \sum_{t=0}^{\infty} \alpha_t^2 &= 0 \label{def:rm_cond_2}
\end{align}
The remaining part of this section will study variations and extensions of the general above method. Concretely, it will be applied to the dynamic programming and reinforcement learning scenario. We notify the reader already that the convergence proofs of these algorithms are beyond the scope of the current work and refer the reader for an overview to \cite{bertsekas1996neuro}, Chapter 4 or \cite{szepesvari2010algorithms}, Chapter 3.


\subsection{Value Approximators}
In this part, we will be looking at approximating the value function $V^{\pi}$. This means that we will be trying to develop methods that output an estimated value for every state given some policy $\pi$. With an estimated version of the value function, we can then apply the generalised policy iteration to obtain an improved policy. The convergence of these methods is discussed in greater detail in \cite{szepesvari2010algorithms, bertsekas1996neuro, Sutton2018}.

\subsubsection{Monte Carlo - Methods}
Monte Carlo methods are designed for computing stochastic integrals such as expectations. We will consider these as a class of stochastic approximations as well due to their conceptual and practical importance. They will also serve as a starting point for the stochastic approximation methods based on Robbins-Monro schemes discussed in the previous section. \\
Starting with the value function given a policy $\pi$.
\begin{align*}
    V^{\pi}(x) &= \mathbb{E} \bigg[\sum_{t=1}^{\infty} \gamma^t R_t | x_0=x\bigg]
\end{align*}
We see that the random variable of interest is $G = \sum_{t=1}^{\infty} \gamma^t R_t$. Given a starting state $x_0=x$ and a policy $\pi$ we can generate trajectories $episode_k=(x_0^k,a_0^k,x_1^k,R_0^k,...,x_T^k,R_{T-1}^k)$ in a finite-horizon MDP. We can use this trajectory to compute $G_k(x) = \sum_{t=0}^{T-1} \gamma^t R_t$. We can then construct a simple sample mean over K trajectories.
\begin{align}
   \tilde{V}^{\pi}_K(x) = \frac{1}{K} \sum_{k=1}^K G_k(x) 
\end{align}
This forms an unbiased estimate of $V^{\pi}(x)$, which is easy to see, since all the $G_k$ are independent and finite and the expectation is a linear operator. Furthermore, the standard deviation decreases as 1/$\sqrt{K}$ by a similar argument. Hence, by the laws of large numbers we are guaranteed the convergence to $V^{\pi}(x)$. If we were to implement this naively and only use every episode for a single state, we can clearly see that it would require many episodes to converge. This gives rise to the first visit Monte Carlo estimate of the value function, which can update the value function for more than one state per episode.
\begin{algorithm}[H]
\setstretch{1}
\SetKwInOut{Input}{Input}\SetKwInOut{Output}{Output}
\Input{Policy $\pi$, Simulator of MDP}
\Output{$V_{\pi}$}
\BlankLine
 $V(x)\leftarrow rand(\mathbb{R})$,   $\forall x\in\mathcal{X}$ \; 
 Returns(x)  $\leftarrow$ emtpy list,   $\forall x\in\mathcal{X}$ \;
 \BlankLine
    \While{True}{
        episode = sample trajectory from simulator following $\pi$\;
        $G \leftarrow 0$\;
        \BlankLine
        \For{$t\leftarrow T-1$ \KwTo $0$ } {
            $G \leftarrow \gamma G + R_t$\;
            \If{$x_t \not\in \{x_0,...,x_{t-1} \}$ }{
                Returns.append(G) \;
                 $V(x)\leftarrow mean(Returns(x))$\;
            }
        }    
    }
 \caption{First Visit Monte Carlo Estimate}
 \label{alg:first_visit_mc}
\end{algorithm}

\noindent In the best case scenario this algorithm generates one trajectory for each state per episode, hence, again this could take a long time to converge. In fact it turns out that removing the first visit condition, $x_t \not\in \{x_0,...,x_{t-1} \}$, also converges to the correct value function. This is not as trivial to see as the samples of the return are no longer independent, for a discussion we refer to \cite{Sutton2018}.\\
We have outlined a Monte Carlo based method for evaluating a policy. Monte Carlo methods, although with theoretical guarantees, sometimes take very long times to converge. Furthermore, if the problem has an infinite time-horizon then Monte Carlo schemes become infeasible. This motivates an alternative approach.

\subsubsection{Temporal Difference - TD(0)}
In this section, we will describe a method that is called temporal difference. We will motivate it using the Robbins-Monro and Monte Carlo method. As in the Monte Carlo setting, we are interested in estimating a value function given a policy $\pi$. Let us start with the Robins-Monro definition considered above (\ref{eq:robbins-monro}). 
\begin{align*}
    x_{t+1} &= (1-\alpha) x_t + \alpha f(x_t,\tilde{z}_t)
\end{align*}
In order to avoid confusion $x_t$ in the above equation is a placeholder for any value that we are trying to estimate. Thus rewriting the $x_t$ in the above equation as $V_t(s)$, $\forall s\in\mathcal{X}$, we obtain:
\begin{align}
    V_{t+1} &= (1-\alpha) V_t + \alpha f(V_t,\tilde{z}_t)   
\end{align}
Now, if we further define the sample $z$ as $z_{t}=(a_{t},R_{t},s_{t+1})$ and the function $f$ as $f(V_{t},z_{t})=R_{t} + \gamma V_{t}(s_{t+1})$. We arrive at:
\begin{align}
    V_{t+1}(s_t) &= (1-\alpha) V_t(s_t) + \alpha \Big[ R_{t} + \gamma V_{t}(s_{t+1}) \Big] \\
    &= V_t(s_t) + \alpha \Big [ R_{t} + \gamma V_{t}(s_{t+1}) - V_t(s_t) \Big ] \label{eq:TD_update}
\end{align}
The last line (\ref{eq:TD_update}), which follows simply by rearranging, is the famous temporal difference update. Also, since we have derived it from the Robin-Monro method, this gives us convergence for free given that $\alpha$ satisfies the RM-conditions (\ref{def:rm_cond_1}) and (\ref{def:rm_cond_2}). The difference term $R_{t} + \gamma V_{t}(s_{t+1}) - V_t(s_t)$ is called the temporal difference error, hence the name.\\
To wrap TD(0) up we will refer the reader to \cite{bertsekas1996neuro}, Chapter 5 and Chapter 6 for an in-depth discussion and only note that TD(0) is a biased estimator with a lower variance than Monte Carlo.

\subsubsection{TD(k) - a unifying view}
Looking at the difference between TD(0) and the Monte Carlo method we will briefly show how they lie on opposite ends of a similar approach. Their difference lies in how long a single trajectory or data-point should be. The natural extension is to consider a method that sits somewhere in-between, i.e. using a k-step look-ahead:
\begin{align}
   V_{t+1}(s_t)  &= V_t(s_t) + \alpha \bigg [ \sum_{l=0}^{k-1} \Big [R_{t+l} + \gamma V_{t}(s_{t+l+1})\Big ] - V_t(s_t) \bigg ]
\end{align}
This formulation makes it apparent that both TD(0) and the Monte Carlo methods are on the different ends of this k-step look-ahead method, with k=0 for TD(0) and k=$\infty$ and $\alpha=1$ for Monte Carlo. The only question remaining is whether these k-step look-ahead also converge to the value function. This is easy to see since again we can use Robbins-Monro by arguing that this time the sample $z_t$ is a trajectory of length k.\\
Now, having discussed that Monte Carlo methods have high variance and are unbiased, while TD(0) has lower variance yet higher bias, this leads us to consider an interpolation between any of these two methods $\tilde{V}=\lambda TD(k_1) + (1-\lambda TD(k_2)$. In fact if we consider an interpolation between all k-step variations and let $\lambda_k = \lambda^k(1-\lambda)$ for $k=0,1,2,...$ then we arrive at an algorithm called TD($\lambda$), we refer the reader to \cite{bertsekas1996neuro}, Chapter 5. $\lambda$ or k only influence the speed of convergence not whether these methods will converge (at least in this particular setting). The final choice of $\lambda$ depends on the problem. For a good list of examples we refer the reader to \cite{Sutton2018}, Chapter 6.
\subsubsection{Summary}
TD($\lambda$) or TD(k) have been developed to approximate the value function. The reason for the approximations were large state and action spaces and more importantly unknown environment dynamics. The TD(k) methods presented in this section are so-called tabular approximation algorithms, as they assign a distinct value for each state. Therefore large state and action spaces would still be problematic. On the other hand, the estimates themselves are obtained without explicitly referring to the environment dynamics. However, ultimately we are interested in the optimal policy, not just the value function. With an estimate of the value function $V^{\pi}$ we can use the generalised policy iteration (Algorithm \ref{alg:generalise_policy_iteration}) to obtain the optimal policy. Looking closely we observe that the generalised policy iteration requires the knowledge of the environment, or more concretely of the transition probabilities. This means that in the more realistic setting of unknown or intractable environment dynamics we need to develop additional methods. Therefore, we can conclude that TD(k) in its standard form is not applicable in the model-free setting.
\subsection{Q-learning}
Q-learning has gained great popularity and considerable success in its applications \cite{mnih2015human}. In this section, we want to investigate this Q-learning algorithm, which nowadays more likely refers to a family of algorithms with various subtle differences. We hope to shed light upon a method that overcomes the short-coming of TD(k), namely the necessity of knowing the model dynamics for the policy update step. Furthermore, we will briefly discuss the extension of the Q-learning algorithm for the non-tabular case and therefore we will arrive at a method that overcomes all mentioned short-comings of exact dynamic programming. Q-learning, among other interesting properties, works under the model-free setting and most interestingly has convergence guarantees to the optimal policy \cite{watkins1992q}, \cite{bertsekas1996neuro}, Chapter 6. Its convergence can be showed using the Robbins-Monro argument in the tabular case and is still being studied for various forms of function approximators. \\
Before we begin describing the original Q-learning algorithm \cite{watkins1992q}, we will briefly mention some of the comparisons of the wider Q-learning approach to other methods. At its core, Q-learning is very similar to temporal difference methods and relies on k-step trajectories and RM conditions. At the same time, Q-learning does not estimate the action-value function under a given policy $Q^{\pi}$, but rather finds the optimal Q-function $Q^{*}$ directly. The direct equivalent to TD estimation for the Q-function is Sarsa \cite{rummery1994line}.
\subsubsection{Tabular Q-learning}
The original Q-learning algorithm \cite{watkins1992q} is based on a tabular representation of both the state and action space and the update rule is given by:
\begin{align}
    Q_t(x_t,a_t) = Q_k(x_t,a_t) + \alpha \Big [ R_t + \gamma \max_{a\in\mathcal{A}}Q_t(x_{t+1},a) - Q_t(x_t,a_t) \Big]
\end{align}
We can easily see that the Robbins-Monro argument is applicable by setting $f(Q_t,z_t)=R_t + \gamma \max_{a\in\mathcal{A}}Q_t(x_{t+1},a)$. The samples $z_t=(R_t, x_{t+1})$ can be collected from an arbitrary policy so long as each state and action is visited infinitely often for full convergence, as is the case for all of these approximation schemes. This particular point will become more important when we are discussing the importance of exploration.
\subsubsection{Q-learning with Function Approximators}
Once we move away from the tabular case various assumptions that helped us with convergence guarantees start failing. Tsitsiklis and Van Roy in \cite{tsitsiklis1997analysis} provide a thorough overview. The main takeaways are as follows. Firstly, with linear function approximators and some technical assumptions on the MDP and approximation such as ergodicity, we are guaranteed convergence in the on-policy case. However, in the case of off-line or non-linear function approximators \cite{tsitsiklis1997analysis}, sections IX and X, these methods are prone to diverge. We refer the reader for a comprehensive overview in Sutton and Barto \cite{Sutton2018}, Chapters 9-12. \\
In this section, we will outline Q-learning using parametrised non-linear function approximators in an off-policy setting. As mentioned above these methods are highly prone to diverge and are to this date poorly understood from a theoretical point of view. Therefore, we will merely outline the key ingredients that allow for non-linear, off policy Q-learning as is suggested empirically. DQN \cite{mnih2015human} is a prominent example thereof. \\
\textbf{Update Rule under function approximators}\\
The general principle under the function approximator does not change. This means that one is still aiming to get closer to the k-step look-ahead target. Hence assuming any parametrisation with parameters $\theta$ the principle is in some sense to minimise:
\begin{align}
    d\Big( Q_{\theta_t}(x_t,a_t),R_t+\gamma \max_{a}Q_{\theta_t}(x_{t+1},a)\Big)
\end{align}
for some distance measured, such that the converged $Q_{\theta_{\infty}}$ is close under max-norm to the optimal solution $Q^{*}$. It turns out that using the least-square distance for d has some theoretical guarantees, especially in the linear approximation case \cite{tsitsiklis1997analysis}.\\
\textbf{Decorrelating Targets} \\
One of the key reasons for divergence of the above method for non-linear and off-policy methods is the correlation between the targets and the online approximator. To overcome this, one introduces an additional function approximator that has the exact functional form of the final approximator. The parameters of this target approximator are updated directly from the final approximator, although on a slower time-scale \cite{Sutton2018}. This results in the following update rule:
\begin{align}
    \theta_{t+1} = arg\min_{\theta_t}\Big( ||R_t+\gamma \max_{a}Q_{\theta'}(x_{t+1},a) - Q_{\theta_t}(x_t,a_t)||_2^2 \Big)
\end{align}
At the same time updating $\theta'$ every T time-steps by setting it equal to $\theta$. These two-timescale methods are known to have some additional convergence guarantees \cite{borkar1997stochastic}.\\
\textbf{Experience Replay} \\
Another way of solving the divergence of the off-policy method is by introducing experience replay \cite{mnih2015human}.  The main idea of experience replay is to gather trajectories $(x_t,a_t,x_{t+1},r_{t+1})$ in the environment to build an array of such experience data points and then to sample these past data points. It turns out that in practice one of the key components to robust convergence comes from experience replay (ER).\\
In summary, we have seen that the Q-learning method provides a way of solving the MDP problem without knowing the underlying dynamics, i.e. in the model-free setting. In addition to this, we have briefly discussed how Q-learning can be done using function approximators and observed that these methods are still poorly understood and are prone to diverge under the interesting scenarios of non-linear function approximators and the off-policy setting. We conclude that establishing a better theoretical understanding presents another valuable direction for future work.

\subsection{Actor Critic}
So far all the methods we have looked at relied on the value function or its relative the state-action value function. However, there is another family of algorithms, which are called policy gradient algorithms. These look to improve the policy function directly without the value function \cite{williams1992simple}. The full scope of these policy gradient methods is vast and probably worth another independent thorough investigation. In this section, we merely want to provide an overview that will be sufficient to develop a Bayesian adaptation of these. \\
Apart from intrinsic motivations for the policy gradient approach, there are three motivations that we will briefly discuss. Firstly, the above-mentioned convergence problems of non-linear, off-policy value approximators lend themselves to the question of how to improve the stability of these systems. It turns out that policy gradient approaches can assist this as the policy gradient theorem demonstrates \cite{Sutton2018}, Chapter 12. Secondly, all the methods discussed for exact or approximate dynamic programming relied on the fact that $\max_a Q(x,a)$ can be solved easily or at all, however, when the action space becomes continuous or high-dimensional this suddenly can turn into a non-trivial problem on its own. Thirdly and perhaps most interestingly, the state-action value approach is designed for the MDP setting, however, many problems do not fully adhere to the Markovian assumption. Using a policy function can lead to non-linear and non-Markovian behaviour and therefore overcome these problems. We will see an example of this with the mountain-car problem in Chapter \ref{chapterlabel5}.  \\
In this section, we will briefly describe the main ideas of policy gradient methods and then we will explain how these can be used in conjunction with value function approximators resulting in the family of methods called actor-critic. The policy function is referred to as actor and the value function is referred to as critic.
\subsubsection{Policy Gradient Methods}
The general idea of policy gradient methods is to find the policy $\pi$ that minimises:
\begin{align}
  \pi^{*} = arg\min_{\pi} \mathbb{E} \Big [R_t\Big]  
\end{align}
Hence, this is a search in the space of all possible policies. Just to give a general intuition that this space is very large let us consider an MDP with a state-space of size X and uniform action space of size A, then there are $A^X$ possible policies. I.e. it is the size of the full breadth-first search tree. \\
Therefore the appropriate methods for this perspective of the problem are of an approximate nature and as the name suggest usually follow some sort of gradient stepping approach. Assuming that the policy is parametrised by $\theta$ the main idea of these approaches lies in finding the gradient:
\begin{align}
   \nabla_{\theta} \mathbb{E} \Big [R_t\Big] \label{eq:policy_gradient}
\end{align}
Computing this directly is near to impossible. This is where the policy gradient theorem comes in \cite{szepesvari2010algorithms}, Chapter 4. The policy gradient theorem states that under various technical assumptions:
\begin{align}
  G(\theta) &= \Big(Q^{\pi_{\theta}}(x,a) - h(x) \Big)\frac{\partial}{\partial \theta} log \pi_{\theta} (a|x) \label{eq:policy_gradient_theorem}
\end{align}
is an unbiased estimate of (\ref{eq:policy_gradient}), where h(x) is an arbitrary bounded function and usually serves the purpose of reducing the variance. A typical choice for h(x) is $V^{\pi}(x)$ which results in the advantage policy gradient method.
\subsubsection{Actor Critic}
In this part we will briefly outline how the actor and critic can work together and therefore one can obtain improved results. In fact, exactly this interaction between actor and critic was one of the factors that led to the success in Alpha Go \cite{silver2016mastering}. Starting with the unbiased estimate of the policy gradient:
\begin{align}
  G(\theta) &= \Big(Q^{\pi_{\theta}}(x,a) - h(x) \Big)\frac{\partial}{\partial \theta} log \pi_{\theta} (a|x) 
\end{align}
we can see that we need estimates of the state value function for this method to work in practice, hence we obtain the gradient that can be used to update the parameters of the actor $\pi_{\theta}$ based on an estimate of the critic $\tilde{Q}_{\psi}^{\pi_{\theta}}$:
\begin{align}
  G(\theta) &= \Big(\tilde{Q}_{\psi}^{\pi_{\theta}}(x,a) - h(x) \Big)\frac{\partial}{\partial \theta} log \pi_{\theta} (a|x) \label{eq:actor_critic_update_rule}
\end{align}
While the critic is updated using the methods developed in the previous section. Exactly this inter-play in updates leads to methods called actor-critic. In this particular adaptation, $\tilde{Q}_{\psi}^{\pi_{\theta}}$ would have to be updated using the on-line version of Q-learning, Sarsa. Such two-timescale algorithms are furhter studied in \cite{borkar1997stochastic} and show promising directions for future work.
\subsection{Summary}
In this section, we have given an overview of the main approximation techniques to exact dynamic programming. These methods are both the foundation of modern state-of-the-art approaches as well as the current state of theoretical foundations. The full analysis of convergence results, conditions and speeds is an interesting study on its own and I hope that more work in this direction will be done by the community.

\section{Exploration vs. Exploitation}
Exploration vs. Exploitation is a very fundamental problem in Reinforcement learning especially more so in real-world scenarios, where someone might be more interested in finite time horizons as opposed to asymptotically optimal solutions. Furthermore, there is even more importance to this topic when considering the approximate methods discussed in the previous section, as they rely on exploring the state and action space exhaustively.

\noindent We will endeavour to give a short overview of the exploration vs. exploitation problem in general and a short overview of various methods to address this issue. Finally, we will briefly remark that on the theory side there was significant work produced by Bellman, Gittins et. al. on the topic of optimal exploration, however, these methods suffer even worse from the "curse of dimensionality`` than the exact dynamic programming algorithms.

\subsection{Importance of Exploration in Approximate Methods}
Exact Dynamic Programming discussed in the first part of this Chapter presented various algorithms, which can all be viewed as special instances of the generalised policy iteration algorithm. We have presented a small proof that convergence can be guaranteed in the finite case rather simply by a contraction mapping argument. In more general settings we have referred to \cite{bertsekas1996neuro} it turns out that some of the key technical conditions for convergence of these methods are the ergodicity of the Markov Decision Process. This is required so that all states are visited sufficiently often. This is exactly the reason why exploration is so important at arriving at an optimal policy or near-optimal solution. 

\subsection{Various Methods of Exploration}
In this section, we will give a short overview of the most prominent methods of exploration. \\
\textbf{$\epsilon$-greedy}\\
The $\epsilon$-greedy method relies on creating trajectories by not only sampling the greedy policy with respect to the Q-function, but acting completely randomly $\epsilon$ amount of the time. In some adaptations, one makes sure that epsilon decreases over time. These methods are perhaps the most widely used in practical RL, however, there are countless examples where $\epsilon$-greedy can lead to near worst-case performance \cite{Sutton2018, osband2017deep}.\\
\textbf{Upper Confidence Bounds (UCB)}\\
This method aims at promoting exploration by introducing pseudo-counts. In particular, the action is selected according to:
\begin{align}
    a_t = arg \max_a \Bigg [Q(x,a) + \sqrt{\frac{log(t)}{N_t(x,a)}}\Bigg ]
\end{align}
Where $N_t(x,a)$ represents the number of times action a was selected in state x. This method has been often shown to perform quite well in practice \cite{Sutton2018}, however, keeping track of $N_t(x,a)$ can become very difficult and it is lacking some stronger theoretical results.\\
\textbf{Optimistic Value Initialisation}\\
Optimisitc Value initialisation is based on its description. The core idea is to initialise the estimates of $Q(x,a)$ to be very large at the very beginning, so that every state is over-estimated and therefore encourages exploration.\\
\textbf{Thompson Sampling}\\
Thompson sampling \cite{thompson1933} is a Bayesian approach that requires uncertainty estimates in the Q-function. Given the full Bayesian treatment the action is chosen according to:
\begin{align}
    a_t = arg \min_a (\tilde{q}(x,a))
\end{align}
where $\tilde{q}$ is sampled from the posterior state-action value function, $\tilde{q} \sim \tilde{Q}_t(x,a)$.

\subsection{Optimal Learning - the solution to exploration}
This part aims to briefly give an overview of what optimal learning is. The main idea of optimal learning lies in taking a fully Bayesian approach on dynamic programming and considering an object called the \textit{information state}. The information state consists of the usual "physical`` state from the state space as well as the Bayesian parameters of the priors. The new transition model now incorporates the effect of information gain or the update to these Bayesian parameters with each action and state observed. To give an example consider a 2-armed Bernoulli Bandit problem with two Beta distributions with parameters $(\alpha_1, \beta_1)$ and $(\alpha_2, \beta_2)$ as priors for each arm respectively. The physical state, in this case, is fixed and can be represented as x=0 without loss of generality. The information state is represented as $(x,\alpha_1, \beta_1, \alpha_2, \beta_2)$. Now, the agent has two actions available, the agent can choose arm one or arm two and for each choice, there are two outcomes each, either success or failure. Therefore an example transition of the information state starting at $(x=0, \alpha_1=1, \beta_1=1, \alpha_2=1, \beta_2=1)$ and the agent picking action one and receiving the success signal results in the new information state of $(x=0, \alpha_1=1+1, \beta_1=1, \alpha_2=1, \beta_2=1)$. At each stage, the agent keeps track of the own reward probabilities and uncertainties thereof resulting even for a 2-armed Bernoulli bandit in an infinite-horizon MDP. It is clear that solving this MDP would result in an optimal path of exploring the two actions given the information that the agent has. However, it is also evident that this is not tractable in any reasonable scenario as even for bandit problems the information state grows exponentially with every arm added.\\
As a final note, it is somewhat surprising that optimal learning is very much untreated by the recent mainstream reinforcement learning literature including seminal works and exhaustive summaries such as \cite{Sutton2018, bertsekas1996neuro, mnih2015human}, perhaps the most recent work on optimal learning is by Powell et. al. (2012) \cite{Powell2012OptimalLP}. In the spirit of approximations of MDPs and exact dynamic programming, we conclude that optimal learning should be explored further in a similar manner.


\chapter{Design, Method and Experiment}
\label{chapterlabel4}

The background and methods for reinforcement learning were discussed in Chapter 2. We have discussed exact dynamic programming, approximate methods thereof, in particular, those that apply in the model-free setting, which is significantly more common and more useful in the real world. In this chapter, we will be presenting our study and contributions with the actual empirical results following in the next Chapter. The theoretical foundation from Chapter 2 serves us to derive a novel theoretical algorithm that combines the actor-critic approach and Bayesian exploration. We empirically evaluate the benefit of the actor-critic (or two-timescale) approach and Bayesian exploration individually on a variety of standard benchmarks as well as a state-of-the-art evaluation benchmark. Furthermore, we will outline our software packages and implementations, and our contribution in this respect is two-fold: Firstly, we are open-sourcing a framework for building and testing arbitrary RL algorithms and provide various in-built environments and models. Secondly, we have implemented a state-of-the-art Bayesian model using accelerated code that allows the usage of GPUs and other similar hardware accelerators. This implementation allows for much larger state and action spaces. We are fully open-sourcing all implementations and making it available under a permissive license. 

\section{Algorithmic Contributions}
Our main algorithmic contribution lies in discussing and proposing the Bayesian actor-critic approach as well as the motivating factors. We also briefly discuss another method that we propose. We call the second method Frequentist Thompson DQN, which is a direct adaptation of \cite{Azizzadenesheli2018} with a frequentist estimate of uncertainty. This method is merely briefly outlined as it is a potential alternative to Bayesian Thompson sampling.
\subsection{Bayesian Actor Critic}
We now present the Bayesian Actor-Critic Method. In Chapter 2 we have seen that the generalised value iteration algorithm with function approximators, i.e. Q-learning can be extended to incorporate another type of learning, namely policy function learning. This, as was shown in \cite{bertsekas1996neuro}, leads to expedited convergence of the algorithms. Together with the performance guarantees and empirical evidence for Thompson sampling, this motivates the Bayesian actor-critic method. 
\subsubsection{Algorithm} 
The main idea of the algorithm lies in two aspects, firstly, maintaining a full Bayesian Q-function with uncertainty estimates, secondly, using a modified version of Thompson sampling together with upper confidence bounds, which I call Bayesian UCB, to pick the action and therefore the exploration trajectories. \\
The inspiration for the modified upper confidence bounds comes from AlphaGo Zero \cite{alphaGoZero}. In their work they pick the action according to $a_t=arg\max_a (Q(x,a) + \frac{P(x,a)}{1+N(x,a)})$, where P(x,a) is given by the policy network and N(x,a) are counts. Naturally, this applies well only when Q(x,a) is rescaled to be within [0,1], which can be accomplished by normalising the rewards of the environment by $R_{max}$. The proposed adaptation to incorporate Thompson sampling is given by: $a_t=arg\max_a (\tilde{q}(x,a) + P(x,a))$, where we get rid of the counts N(x,a) as they might be difficult to keep track of and we sample $\tilde{q} \sim Q$ from the posterior.
\begin{algorithm}[H]
\setstretch{1}
\SetKwInOut{Input}{Input}\SetKwInOut{Output}{Output}
\Input{Simulator of MDP, BatchSize, PolicyUpdateT, TargetUpdateT}
\Output{$V_{\pi}$}
\BlankLine
 $\theta \leftarrow rand()$ \; 
  $\theta' \leftarrow rand()$ \;
 $\psi\leftarrow rand()$ \;
 ReplayBuffer $\leftarrow$ emptyList \;
 \BlankLine
    \While{True}{
        $x_{t+1},R_t,a_{t+1} \leftarrow$ Simulator.act(ThompsonUCB) \;
        ReplayBuffer.append($x_t, a_t, x_{t+1}, R_t, a_{t+1}$) \;
        \BlankLine
         \If {ReplayBuffer.size() $>$ BatchSize}{
            batch $\leftarrow$ ReplayBuffer.sample(BatchSize)\;
            $\theta \leftarrow$ BayesianPosteriorUpdate$\Big ( R_t + \gamma Q_{\theta'}(x_{t+1},a_{t+1}) , Q_{\theta}(x_t,a_t) \Big )$\;
            \If {t mod PolicyUpdateT==0}{
            $\psi \leftarrow \Big(Q_{\theta}(x_t,a_t) - Q_{\theta}(x_t,\pi_{\psi}(x_t)) \Big)\nabla_{\psi} log \pi_{\psi} (a|x)$\;
            }
            \If {t mod TargetUpdateT==0}{
            $\theta' \leftarrow \theta $\;
            }
         }
    }
 \caption{Bayesian Actor Critic}
 \label{alg:bac}
\end{algorithm}
\subsection{Frequentist Thompson DQN}
BDQN\cite{Azizzadenesheli2018} relies on a neural network as a feature map from the state representation and computes the action value function by solving a Bayesian linear regression problem. In our method the idea is to replace the posterior update of the regression step, for computational speed, with a MLE estimate of both the mean and covariance. The policy follows a Thompson sampling approach very akin to the one described in the paper. This method was constructed to potentially serve the validation of Thompson sampling even without proper uncertainty estimates. As, we will see in the Analysis Chapter, however, this method, although it manages to learn, does not perform very well.

\section{Empirical Study}
In this work, we conduct two empirical investigations. On one hand, we investigate bandit problems and study the ability of a variety of approaches to explore and find the correct action. On the other hand, we build upon the state of the art evaluation suite \textbf{bsuite} \cite{osb2019behaviour} released by Google's research group Deepmind. This evaluation is carefully designed to probe a variety of skills of a given algorithm. In total we ran over 25 different experiments each over 20 runs for 5 different agents for bsuite, this resulted in a total of 2500 trained agents. This was only possible by using optimised code and profiling the packages. The code is fully open-source\footnote{\url{https://github.com/ai-nikolai/bsuite} \& \url{https://github.com/ai-nikolai/barl}}

\subsection{Bandit Exploration}
In the bandit setting, we try to understand the agent's ability to find the optimal action in a minimal number of steps. We focused on comparing two standard frequentist approaches vs. Thompson sampling. Furthermore, we compare the agents in different settings of the bandit problem. We vary the noise that the environment has, as well as how distinguishable the various reward signals are. This distinguishability is measured by the L2 distance, with higher Euclidean distance leading to higher distinguishability. \\
\textbf{Environment}\\
We use a k-armed gaussian bandit. This means that the reward for arm $i$ is generated by an independent normal with mean $\mu_i$ and variance $\sigma_i^2$. \\
\textbf{Agents}
\begin{enumerate}
    \item $\epsilon$-greedy agent with MLE estimates of the mean using a Gaussian environment model. 
    \item Optimistic value initialisation with greedy actions and MLE estimate of the mean using a Gaussian environment model.
    \item Thompson sampling with Gaussian environment model and exact posterior updates.
\end{enumerate}
This analysis gives a small insight into the efficiency benefit that uncertainty estimates provide and therefore motivates the usage of such uncertainty estimates provided by Bayesian methods in other scenarios. 

\subsection{State-of-the-art Agent Behaviour Suite}
Decision making agents need to perform well in a variety of tasks. Generalisation, memorisation, exploration, noise, scaling, long-term planning etc. Most real-world applications have a combination of all of these and more distinct problems that an agent needs to solve. A long-standing problem in reinforcement learning research was the ability to compare and assess the agents on all of these individually. Google's Deepmind endeavours to overcome this by providing a general sandbox environment where researchers can test their hypotheses. Therefore, for the purpose of this research, we have chosen to use this framework and extended it to compare our use-cases and algorithms. 

\section{Software}
\subsection{BARL - Framework}
BARL, short for Bayesian Approximate RL, is a framework specifically developed for this thesis and the future work intended by it. This framework consists of three core building blocks. Agents, Estimators and Environments. These are augmented with a variety of auxiliary tools for running simulations, plotting and profiling. This code was profiled to allow faster experimentation. \\
\textbf{Agents}\\
Agents form the general design of an agent, i.e. whether to have experience replay, whether to act greedily, randomly or in some other manner, whether to update every time-step in an online fashion or sample experience replay batches and update in an offline manner, etc. These agents, however, are independent of the estimators and function approximators that they are using. \\
\textbf{Estimators}\\
This part of the package is designed to allow for custom estimators and function approximators, such as Bayesian neural networks, Gaussian Processes or Recurrent Neural Networks \cite{hochreiter1997long}. \\
\textbf{Environments}\\
The package allows for extensibility on the environment side, i.e. it provides an abstract class that allows to build and add additional environments.\\
\textbf{Utility} \\
Finally, the package also provides a variety of utility functions. It allows for simplified access to plotting and running agents and environments by wrapping them as simulations. 

\subsection{Bsuite Extensions}
Bsuite is a rather novel framework that is not a Google product and therefore needs and allows for a lot of customisation. Our main contribution to the package are the following:
\begin{enumerate}
    \item Tensorflow 2.0 implementation of DQN for increased speed-up.
    \item Tensorflow and Tensorflow-Probability implementation of a BDQN variation using MLE updates as opposed to the exact posterior for significant speedup on accelerated hardware.
\end{enumerate}

\section{Summary}
The methods described and presented in this Chapter follow the general agenda of this work on providing motivation and foundation for applicable reinforcement learning. In particular, work that is motivated and grounded in theory of approximate dynamic programming, that allows for large scale state and action spaces with unknown dynamics and that can become applicable in the real world by having increased data-efficiency and good exploration capabilities. \\
In summary, we have presented a novel algorithm as an extension of the actor-critic approximate dynamic programming method and Thompson sampling. We motivated it by the two-timescale approximation guarantees discussed in \cite{bertsekas1996neuro}. We then outlined the empirical setup for validating the survey of ideas and methods presented in this work. Finally, we outlined briefly the software that is accompanying this work.\\

\chapter{Analysis}
\label{chapterlabel5}
The work presented so far went from motivating examples of state-of-the-art approaches to the theoretical foundation of these in exact dynamic programming as well as approximations thereof. We then further discussed the importance of exploration and exploitation. All of this serves as a survey of reinforcement learning from the perspective of approximate dynamic programming as well as a foundation for future work into real-world applicable reinforcement learning.\\
The main theoretical method presented in the previous Chapter expresses some of the directions that our survey suggests are worthy of exploration, namely actor-critic methods as well as Thompson sampling. In this chapter, we provide some results that support this claim. In particular, we focus our analysis on exploration vs. exploitation in a variety of scenarios, as well as on the analysis of the actor-critic method. We conducted these studies in the spirit of ablation studies and therefore compare these two directions against a strong deep q-learning baseline, which hat neither Bayesian exploration nor the actor-critic approach. We also show an analysis of the data-efficiency of Thompson sampling in simple environments.

\section{Exploration in Bandits}
This experiment is aimed at comparing exploration methods in the k-armed bandit setting as outlined in the previous chapter. We ran the experiment with random initialisation of noise and reward means. We further compared the effectiveness of three algorithms: Optimistic Value Initialisation, $\epsilon$-greedy and Thompson sampling. We run these for 300 time-steps for 50 runs and compare them to the best possible action and a random baseline.\\

\begin{figure}[h!]
    \centering
    \includegraphics[width=0.7\textwidth]{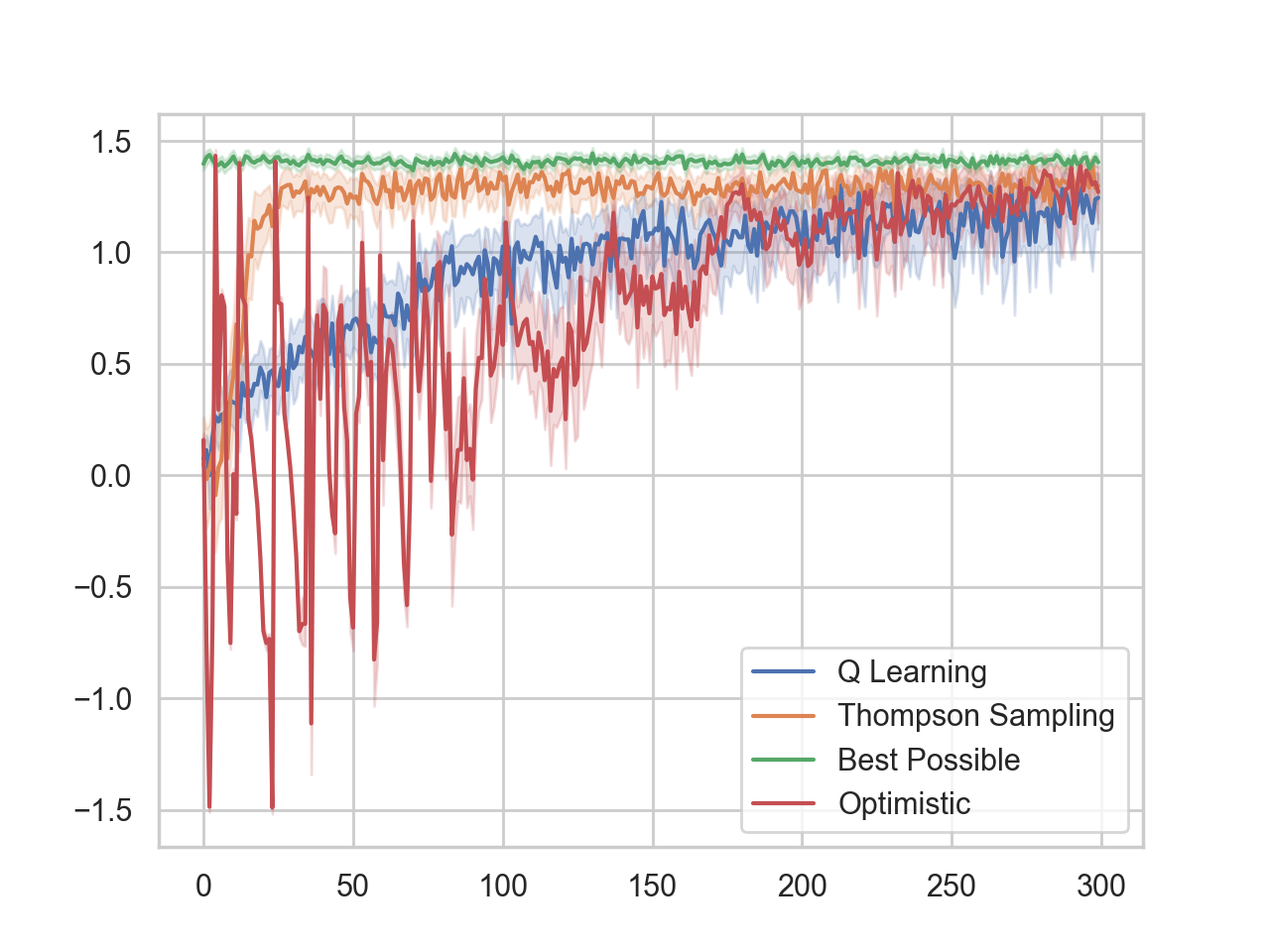}
    \caption{Reward per time-step in a 12 armed Bandit over 50 runs - without random baseline}
    \label{fig:barl_1}
\end{figure}

\begin{figure}[h!]
    \centering
    \includegraphics[width=0.7\textwidth]{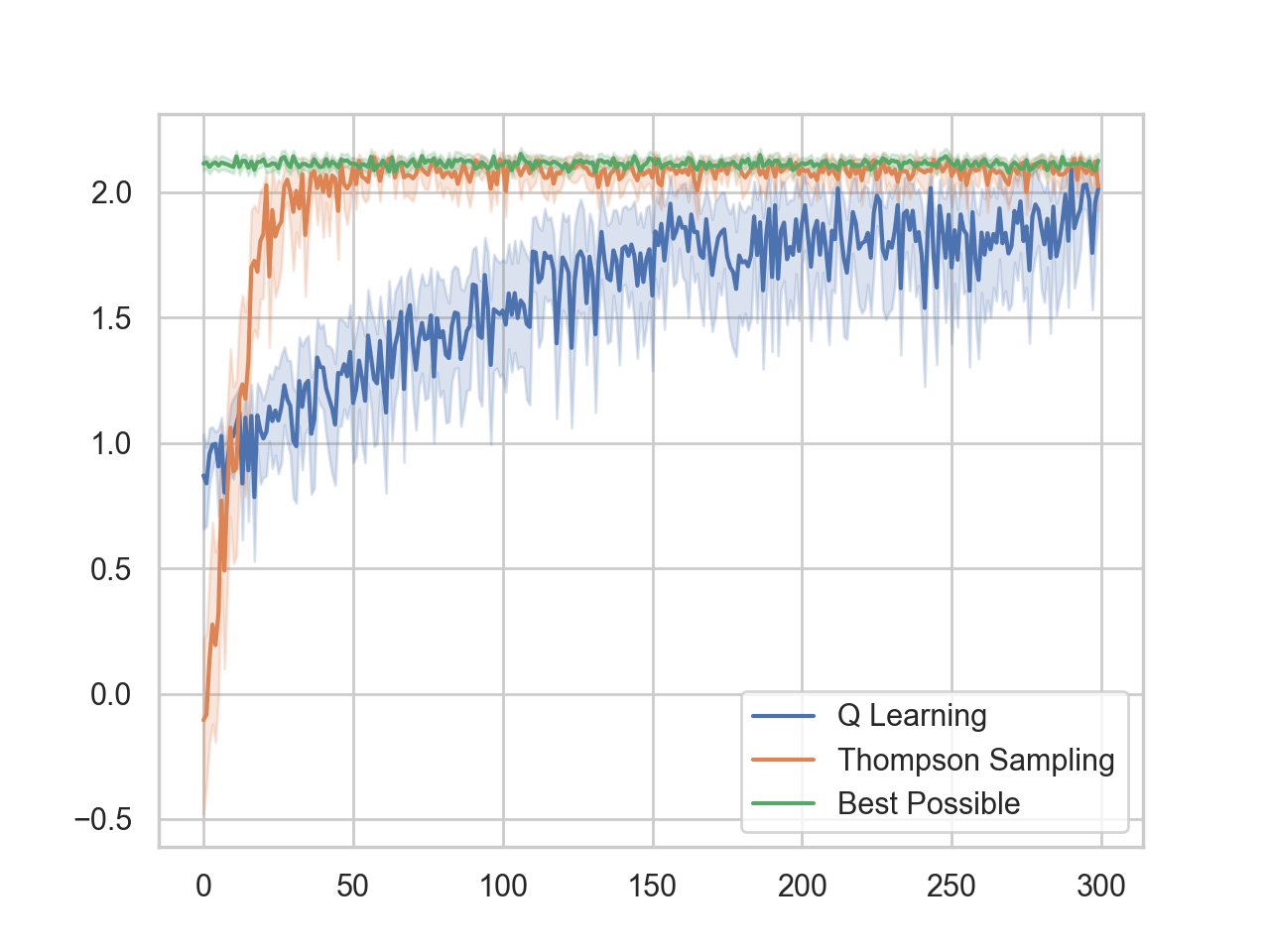}
    \caption{Reward per time-step in a 12 armed Bandit over 50 runs - without optimistic baseline}
    \label{fig:barl_2}
\end{figure}

\begin{figure}[h!]
    \centering
    \includegraphics[width=0.7\textwidth]{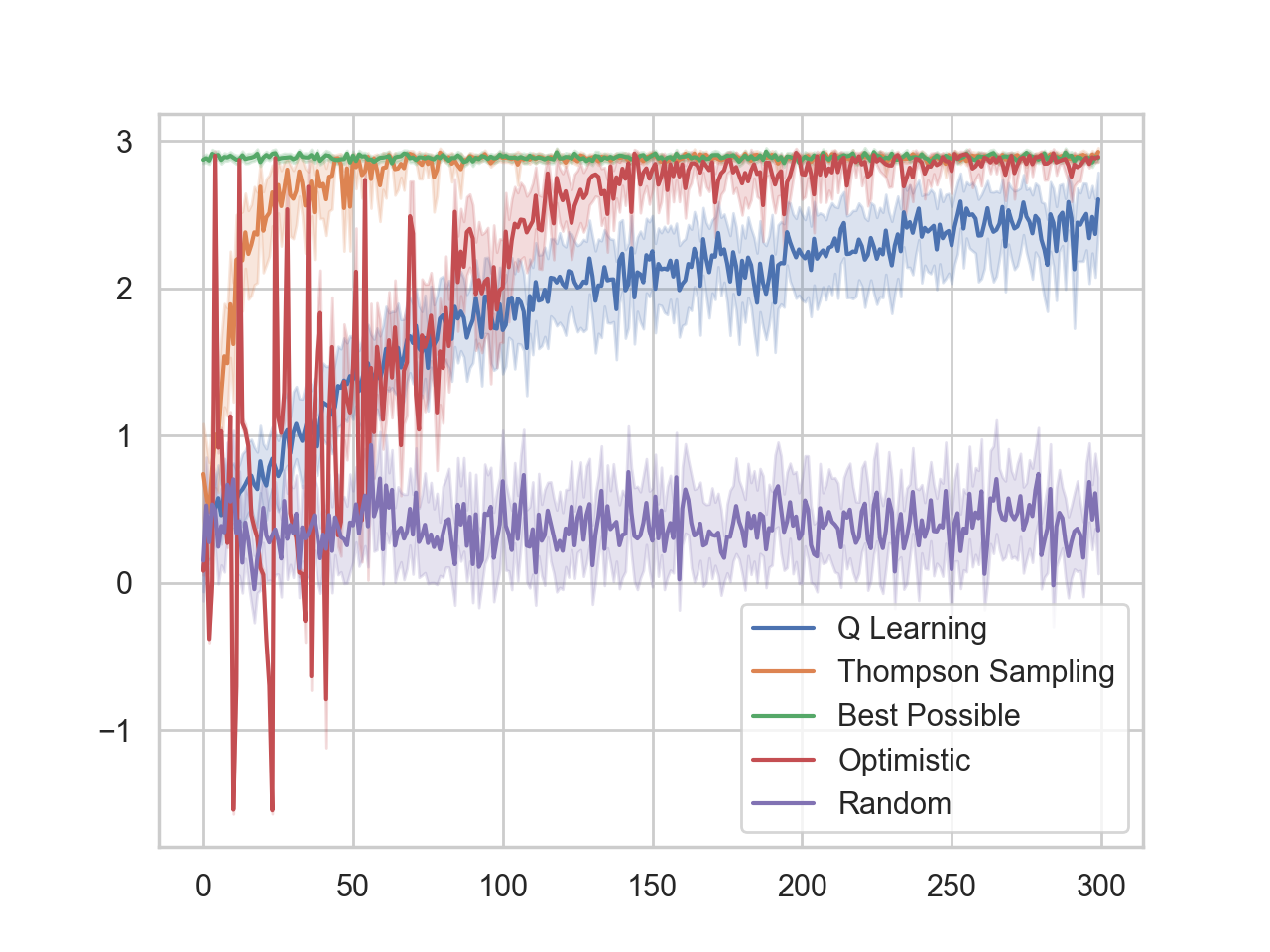}
    \caption{Reward per time-step in a 12 armed Bandit over 50 runs - all agents}
    \label{fig:barl_3}
\end{figure}

\begin{table}[h!]
\centering
\begin{tabular}{ |c|c|c| } 
 \hline
    Name & Mean Reward & Std \\ 
 \hline
 $\epsilon$-greedy & 569.4271 & 58.3278 \\ 
 Optimistic & 672.1069 & 2.2556 \\ 
 Thompson & 833.2319 & 4.3487 \\ 
  Best & 866.5053 & 1.5431 \\
 Random & 115.3462 & 20.7789 \\
 \hline
\end{tabular}
\caption{Bandit Exploration in a random 12 armed bandit over 50 runs.}
\label{table:barl_1}
\end{table}

\noindent In figures \ref{fig:barl_1}, \ref{fig:barl_2}, \ref{fig:barl_3} and table \ref{table:barl_1} we can see that that Thompson sampling is the clear winner among these exploration techniques, also, we can see that compared to the $\epsilon$-greedy approach it performs in a significantly more stable way with a standard deviation of only 4. Another observation that we can make is that the Optimistic approach is very erratic (cf. figure \ref{fig:barl_1}). This behaviour is expected and arises from the fact that after the latest k-1 observations in a k-armed bandit the worst action will be picked with a very high probability. To conclude the Bandit section, we will merely say that as expected and suggested by the literature Thompson sampling indeed performs better in terms of data efficiency and final result.
\section{Agent Behaviour Suite Results}
This section presents the main empirical validation of the ideas presented in this work. In general, we will be discussing three scenarios in more detail that are aimed at showing different parts of importance: Thompson sampling being able to extend to MDPs as a data-efficient method, effectiveness of the actor-critic method in environments that likely violate full Markovian assumptions and the necessity of advanced exploration methods in complicated environments. We also discuss the overall performance and observations of three agents DQN, Bayesian DQN \cite{osband2017deep} and the advantage actor-critic (A2C). In order to analyse this we have trained a vanilla deep q-network with experience replay following many of the suggestions from the original and seminal work on DQNs \cite{mnih2015human}, our frequentist adapted version of BDQN (Frequentist Thompson DQN) \cite{Azizzadenesheli2018}, the full Bayesian Deep Q learning from Osband et. al. \cite{osband2017deep} and finally a vanilla actor-critic algorithm. We trained and evaluated these on 25 environments from bsuite \cite{osb2019behaviour} with over 20 runs for each. The data for the three in-depth experiments is in Appendix \ref{appendixlabel2}.

\subsection{A standard Problem}
Cartpole \cite{barto1983neuronlike} is perhaps the "Hello World" of reinforcement learning in MDPs. The problem description is rather simple: control a simulated cart so that the pole on top of the cart does not fall. The environment is represented by the speed of the cart and the angle of the pole. If the pole tips more than 15 degrees, the cart is 2.5 units away from the centre or 1000 time-steps passed the episode finishes. The agent controls a discrete force of +1 or -1 exerted on the cart and receives a reward of +1 for every time-step that the episode did not finish. Hence, the optimal performance that one can achieve is a score of 1001 per episode.

\begin{table}[h!]
\centering
\begin{tabular}{ |c|c|c|c|c|c|c|c| } 
 \hline
name&steps&episode&total return&ep. len&ep. return&raw return&best episode\\
\hline
BootDQN&9989&70&9920.0&86&85.0&9920.0&1001.0\\
DQN&34463&80&34384.0&622&621.0&34384.0&1001.0\\ 
Freq.BDQN&51510&1000&50510.0&268&267.0&50510.0&595.0\\
A2C&53754&300&53455.0&178&177.0&53455.0&1001.0\\
 \hline
\end{tabular}
\caption{Comparison of Cart-pole environment with four agents}
\label{table:cartpole_all}
\end{table}

\noindent Table \ref{table:cartpole_all} serves to illustrate when the respective Agents first reached the maximal episode length of 1000. We see a significant benefit of using Thompson sampling by the BootDQN over all other methods. Similarly, we see that in this classical control problem DQN performs very well. Finally, we can also observe that the frequentist Thompson sampling failed to learn to balance the pole. I believe that this is due to lack of hyper-parameter tuning. However, this is also a clear indicator that Thompson sampling without proper uncertainty tracking does not offer a robust method.

\subsection{Deriving non-linear policies}
Mountain car \cite{moore1990efficient} is a problem that is often referred to as requiring memorisation \cite{brockman2016openai}. What this means is that the agent needs to remember past actions and learn to create plans over longer periods of time. The problem itself is to move a car along a 1-D path that represents a valley between two mountains. The car itself is not strong enough to make the journey up in one attempt, therefore, the solution is to swing the car there and back to build up momentum. The agent controls whether the car moves forward or backwards and receives negative rewards at every time-step that the car has not reached the top.
\begin{table}[h!]
\centering
\begin{tabular}{ |c|c|c|c|c|c|c| } 
 \hline
name&steps&episode&total return&episode len&episode return&raw return\\
\hline
BootDQN&39884&1000&-39884.0&42&-42.0&-39884.0\\
DQN&50901&1000&-50901.0&37&-37.0&-50901.0\\
Freq.BDQN&516077&1000&-516077.0&41&-41.0&-516077.0\\
A2C&33735&1000&-33735.0&30&-30.0&-33735.0\\
 \hline
\end{tabular}
\caption{Comparison of Mountain-car environment with 4 agents}
\label{table:mountaincar_all}
\end{table}
Table \ref{table:mountaincar_all} serves to demonstrate which agent struggled the most with this example. This is represented by the total return given the same number of episodes. Interestingly enough the performance of the actor-critic algorithm shows the strongest performance by being able to reach the top of the mountain in the fewest steps on average over 1000 episodes. There is a 25\% drop in performance of the runner up, Bayesian DQN and 40\% concerning the vanilla DQN. The frequentist BDQN again performs worst by a huge margin. This simple experiment shows promising results in the intuition that policy gradient-based methods provide and therefore confirms that this is a valid direction for further investigation.

\subsection{Exploring long-term strategies with sparse rewards}
Deepsea \cite{osband2017deep} is an environment that is specifically designed to test the agent's ability for long term exploration. The environment is a conceptual 2-D grid where the agent can either move down-left or down-right. The starting position of the agent is in the top left corner and the only positive reward that the agent receives is by always following down-right. Therefore, a naive exploration policy is bound to have an exploration length of $2^N$, where N represents the size of the world.\\
\begin{table}[h!]
\centering
\begin{tabular}{ |c|c|c|c|c|c|c|c| } 
 \hline
name&steps&ep.&ttl. ret&ep. len&ep. return&ttl. bad ep.&denoised ret\\
\hline
A2C&100000&10000&-9.0830&10&-0.0030&10000&0\\
DQN&100000&10000&-20.8227&10&-0.0011&10000&0\\
Boot&100000&10000&9731.7109&10&0.9900&169&9831.0\\
 \hline
\end{tabular}
\caption{Comparison of Deep sea environment with 3 agents}
\label{table:deep_sea_all}
\end{table}
In table \ref{table:deep_sea_all} we are comparing DQN, Boot DQN and A2C. All agents were trained in an environment with a world size of N=10, the salient feature is "ttl. ret", which is the total return. Boot DQN significantly outperforms its peers and is the only agent that solves the problem. The other interesting observation is that again A2C achieves a better performance in a task that potentially requires longer-term planning than its value-based counter-part DQN. In this experiment, we are not including frequentist BDQN as it was again strongly under-performing.

\subsection{Overall results}
So far we have compared four agents on three specific tasks. The results were promising with respect to the claims made in this work. In particular, the results suggested so far that Thompson sampling indeed allows for increased exploration and data-efficiency and that the actor-critic method allows for non-linear policies and long-term planning. However, whether this holds in general and across a wider variety of tasks was not yet shown. Hence, we also present a comparison of actor-critic, boot DQN and DQN across the whole set of 25 tasks in bsuite \cite{osb2019behaviour}. \\
\begin{figure}[h!]
    \centering
    \includegraphics[width=0.9\textwidth]{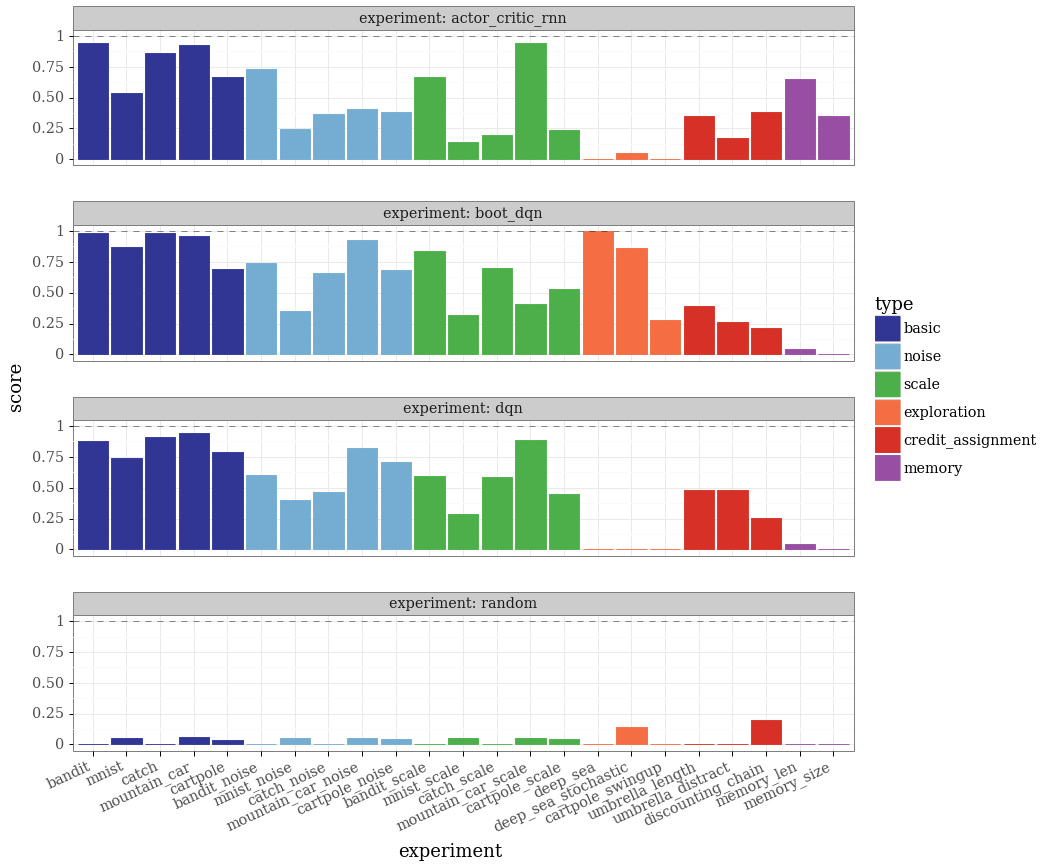}
    \caption{Comparison of three agents across all 25 tasks. }
    \label{fig:bar_plot}
\end{figure}
In figure \ref{fig:bar_plot} we can clearly see that on the basic, noise, scale and credit assignment tasks all three agents perform relatively similar. The significant differences come from the exploration tasks and the "memory" tasks. \\
\textbf{Exploration}\\
Boot DQN, which is based on Thompson sampling performs consistently well on the exploration tasks and, in fact, is the only agent that manages to solve these tasks. These tasks require multi-step action taking and have long event horizons. This means that the Thompson sampling method, which was designed for bandit problems in its original form, seems to generalise across to MDPs. At the same time, Thompson sampling, in this particular set-up, does not allow for the solution of tasks that requires memory and non-Markovian planning. \\
\textbf{Long Term Planning - Memory}\\
The only agent that manages to out-perform the random baseline in the memory tasks is an actor-critic method. This falls in-line with the arguments that we have been putting forward and is encouraging for our theoretical proposal. On a final note, it remains to be seen what precisely makes policy gradient methods and actor-critic methods perform better at such tasks and how the Bayesian perspective can fully utilise it.
\section{Error Analysis}
Apart from the strong suggestions and validation of our ideas, there are some shortcomings and some error analysis that we will briefly discuss. Firstly, the frequentist Thompson DQN is not optimised for hyper-parameters as this would have been too computationally expensive. Hence, it is not fully clear whether full Bayesian uncertainty tracking is necessary. Similarly, it was not possible to replicate the original BDQN paper \cite{Azizzadenesheli2018}, which also relies on Bayesian Thompson sampling, and therefore it remains to be seen whether Thompson sampling truly generalises across methods. Lastly, even if all the claims in this analysis section turn out to be true, it still remains to see whether the combination of actor-critic and Bayesian Thompson sampling indeed has the promised performance gain.

\section{Conclusion}
In this chapter, we have presented empirical results on exploration and on convergence speed comparisons of the value-based approximation schemes vs. the actor-critic based approximation scheme. These experiments were made with repeated tries and in a variety of environments. The results show strong support for Bayesian exploration and actor-critic methods reinforcing the Bayesian actor-critic method proposed in the previous chapter. 

\chapter{Conclusion \& Outlook}
\label{chapterlabel6}
This work's main objective was to contribute to the field of sequential decision making and in particular their application to the real world. This field is very large in size and therefore the concrete steps for this higher-level objective were to provide a survey of both the theoretical foundation of sequential decision making, as well as to provide an overview of the state-of-the-art of this highly varied, fast-changing and rich field. These theoretical foundations have been used to motivate existing methods as well as a new promising direction of algorithms and future work. In particular, the emphasis was laid on exploration in reinforcement learning and the actor-critic method for approximating exact dynamic programming. These aspects of approximate dynamic programming were then analysed empirically and showed promising results. \\
Concretely, we have seen that keeping track of the uncertainty in the state-action value function, together with Thompson sampling, leads to increased data-efficiency and sometimes is the only feasible way of solving the problem. We have observed that the actor-critic algorithm also performs increasingly efficiently in comparison to the standard, yet state-of-the-art, deep q-learning. Finally, we have presented software that was designed and tested for hardware-accelerated devices and allows not only replication of these results but also facilitates future work. \\
The final objective of this work was to lay a foundation for a deep dive into reinforcement learning - such as a contemporary doctoral degree on the topic of sequential decision making. I believe that this has been accomplished, as an overview of the past, the present and the tools of this field have been presented analysed and used. To conclude, some of the future outlook based on this work and beyond will be discussed.

\section{Outlook}
This work has presented the actor-critic family of algorithms and their adaptation to Thompson sampling under Bayesian uncertainty tracking. Due to a variety of limitations in computation available and the vast field of exploration, it was not possible to empirically validate this method. Hence, this is the most natural next stepping stone from this work. Another direction that needs more validation is whether Bayesian neural networks are indeed important for these methods to work or whether alternative Bayesian methods can be even better adapted for this task. The exploration of frequentist uncertainty estimates, as a way of speeding up computation, also needs to be explored further in the context of reinforcement learning. On the conceptual side of things, a proper exploration of convergence guarantees (or lack thereof) of non-linear function approximators for TD based methods is outstanding in this field. Finally, the software developed for this work has a lot of room for improvement and contributions. The software has the potential to facilitate research for many parties and therefore is worth pursuing further.
\subsection{Frontiers}
Apart from the core ideas presented in this work, some peripheral ideas have been mentioned and touched upon. Perhaps the most significant and interesting are: \\
\textbf{Curiosity - or mixed model-based and free RL}:\\
Open AI's work on curiosity-driven exploration is fascinating. Since the agents learn to solve so many tasks without ever being given the true reward signal. The underlying principle of this curiosity-driven exploration is by learning to predict the environment and exploring those parts of it where the prediction is the weakest. Thus it could be argued that curiosity-driven RL is model-based RL, where the model is being learned. The interesting question that can be posed here is how can the exploration methods and convergence methods described and motivated by the theory presented in this work be joined with the curiosity-driven approach.
\textbf{Approximate Optimal Learning}:\\
Optimal learning is potentially the most overlooked aspect of optimal control and sequential decision making in our current research community. Even exhaustive summaries such as Sutton's \cite{Sutton2018} mention optimal learning only in the passing. Naturally, exact optimal learning is far from being tractable. At the same time, exact vanilla dynamic programming is not tractable either. Hence, the natural question that arises is, how can optimal learning be effectively approximated. This question has potentially a link to the frequentist way of tracking uncertainty in reinforcement learning.\\
\textbf{Multi-agent systems}: \\
As a final frontier, many questions lie, of course, in the field of multi-agent learning. Multi-agent RL is still a widely unsolved and untapped field and I am personally looking forward to seeing this field grow into something more mature, as we have seen with reinforcement learning itself.\\

\addcontentsline{toc}{chapter}{Appendices}

\appendix
\chapter{Bellman Optimality Operator is a Contraction Mapping}
\label{appendixlabel1}
Let us proof that the Bellman Optimality Operator defined in equation (\ref{eq:optim_value_equation}) is indeed a contraction mapping on V.
\begin{align}
   &&&||(T^{*}V)(x) - (T^{*}V')(x)||_{\infty} =\\
   &=& &\max_{x\in\mathcal{X}}\Bigg\{\max_{a\in\mathcal{A}} \bigg\{  r(x,a) + \gamma \sum_{x_1\in\mathcal{X}}p(x_1|x,a) V(x_{1}) \bigg\} \nonumber \\
    &&-&\max_{a\in\mathcal{A}} \bigg\{  r(x,a) + \gamma \sum_{x_1\in\mathcal{X}}p(x_1|x,a) V'(x_{1}) \bigg\} \Bigg\}\\
    &\leq& &\max_{a\in\mathcal{A},\, x\in\mathcal{X}} \bigg\{ \cancel{ r(x,a)} + \gamma \sum_{x_1\in\mathcal{X}}p(x_1|x,a) V(x_{1}) \nonumber \\
    &&-& \cancel{r(x,a)} +\gamma \sum_{x_1\in\mathcal{X}}p(x_1|x,a) V'(x_{1}) \bigg\} \\
    &=& \gamma& \max_{a\in\mathcal{A},\, x\in\mathcal{X}} \bigg\{ \sum_{x_1\in\mathcal{X}}p(x_1|x,a) V(x_{1})  - 
    \sum_{x_1\in\mathcal{X}}p(x_1|x,a) V'(x_{1}) \bigg\} \\
    &=& \gamma &\max_{a\in\mathcal{A},\, x\in\mathcal{X}} \bigg\{ \sum_{x_1\in\mathcal{X}} p(x_1|x,a) \Big(V(x_{1})  -  V'(x_{1}) \Big) \bigg\} \\
    &\leq& \gamma & \max_{a\in\mathcal{A},\, x\in\mathcal{X}} \bigg\{  \sum_{x_1\in\mathcal{X}}\bigg[ p(x_1|x,a) \max_{x'\in\mathcal{X}}  \big\{ V(x')  -  V'(x') \big\}   \bigg] \bigg\}\\
    &=& \gamma & \max_{a\in\mathcal{A},\, x\in\mathcal{X}} \bigg\{ \sum_{x_1\in\mathcal{X}}p(x_1|x,a) \bigg\} \; ||V  -  V'||_{\infty}\\ 
    &=& \gamma &  ||V  -  V'||_{\infty}
\end{align}
For $0<\gamma<1$ we see that T is indeed a contraction mapping, this concludes the proof. \\
\null \hfill $\square$

\chapter{Bsuite Data}
\label{appendixlabel2}
Here we are attaching some of the results from the experiments:
\begin{table}[h!]
\centering
\begin{tabular}{ |c|c|c|c|c|c|c| } 
 \hline
steps&episode&total return&episode len&episode return&raw return&best episode\\
\hline
29&1&28.0&29&28.0&28.0&28.0\\
57&2&55.0&28&27.0&55.0&28.0\\
85&3&82.0&28&27.0&82.0&28.0\\
115&4&111.0&30&29.0&111.0&29.0\\
144&5&139.0&29&28.0&139.0&29.0\\
173&6&167.0&29&28.0&167.0&29.0\\
202&7&195.0&29&28.0&195.0&29.0\\
233&8&225.0&31&30.0&225.0&30.0\\
262&9&253.0&29&28.0&253.0&30.0\\
291&10&281.0&29&28.0&281.0&30.0\\
350&12&338.0&30&29.0&338.0&30.0\\
407&14&393.0&29&28.0&393.0&30.0\\
494&17&477.0&28&27.0&477.0&30.0\\
582&20&562.0&28&27.0&562.0&30.0\\
731&25&706.0&30&29.0&706.0&30.0\\
876&30&846.0&29&28.0&846.0&30.0\\
1170&40&1130.0&28&27.0&1130.0&30.0\\
1468&50&1418.0&31&30.0&1418.0&30.0\\
1760&60&1700.0&29&28.0&1700.0&30.0\\
2058&70&1988.0&30&29.0&1988.0&30.0\\
2381&80&2301.0&36&35.0&2301.0&35.0\\
2735&90&2645.0&33&32.0&2645.0&50.0\\
3088&100&2988.0&32&31.0&2988.0&50.0\\
3854&120&3734.0&35&34.0&3734.0&50.0\\
4607&140&4467.0&35&34.0&4467.0&51.0\\
5767&170&5597.0&47&46.0&5597.0&52.0\\
6862&200&6662.0&28&27.0&6662.0&52.0\\
8773&250&8523.0&38&37.0&8523.0&52.0\\
10555&300&10255.0&33&32.0&10255.0&52.0\\
13972&400&13572.0&32&31.0&13572.0&52.0\\
17530&500&17030.0&36&35.0&17030.0&53.0\\
21391&600&20791.0&36&35.0&20791.0&69.0\\
25759&700&25059.0&70&69.0&25059.0&117.0\\
37838&800&37038.0&41&40.0&37038.0&595.0\\
43067&900&42167.0&67&66.0&42167.0&595.0\\
51510&1000&50510.0&268&267.0&50510.0&595.0\\
 \hline
\end{tabular}
\caption{Frequentist BDQN on Cartpole}
\label{table:bdqn_cartpole}
\end{table}
\begin{table}[h!]
\centering
\begin{tabular}{ |c|c|c|c|c|c|c| } 
 \hline
steps&episode&total return&episode len&episode return&raw return&best episode\\
\hline
83&1&82.0&83&82.0&82.0&82.0\\ 
208&2&206.0&125&124.0&206.0&124.0\\ 
287&3&284.0&79&78.0&284.0&124.0\\ 
373&4&369.0&86&85.0&369.0&124.0\\ 
426&5&421.0&53&52.0&421.0&124.0\\ 
587&6&581.0&161&160.0&581.0&160.0\\ 
668&7&661.0&81&80.0&661.0&160.0\\ 
719&8&711.0&51&50.0&711.0&160.0\\ 
798&9&789.0&79&78.0&789.0&160.0\\ 
971&10&961.0&173&172.0&961.0&172.0\\ 
1215&12&1203.0&111&110.0&1203.0&172.0\\ 
1371&14&1357.0&95&94.0&1357.0&172.0\\ 
1780&17&1763.0&84&83.0&1763.0&212.0\\ 
2136&20&2116.0&131&130.0&2116.0&212.0\\ 
2751&25&2726.0&131&130.0&2726.0&212.0\\ 
3599&30&3569.0&157&156.0&3569.0&212.0\\ 
8788&40&8748.0&628&627.0&8748.0&861.0\\ 
16510&50&16460.0&733&732.0&16460.0&882.0\\ 
24754&60&24694.0&478&477.0&24694.0&906.0\\ 
28488&70&28418.0&316&315.0&28418.0&906.0\\ 
34463&80&34384.0&622&621.0&34384.0&1001.0\\ 
40727&90&40643.0&189&188.0&40643.0&1001.0\\ 
46167&100&46076.0&614&613.0&46076.0&1001.0\\ 
54417&120&54306.0&714&713.0&54306.0&1001.0\\ 
64573&140&64443.0&555&554.0&64443.0&1001.0\\ 
89209&170&89066.0&1001&1001.0&89066.0&1001.0\\ 
109211&200&109048.0&478&477.0&109048.0&1001.0\\ 
140929&250&140735.0&359&358.0&140735.0&1001.0\\ 
176244&300&176020.0&280&279.0&176020.0&1001.0\\ 
263920&400&263663.0&1001&1001.0&263663.0&1001.0\\ 
349184&500&348900.0&245&244.0&348900.0&1001.0\\ 
438149&600&437838.0&1001&1001.0&437838.0&1001.0\\ 
522124&700&521783.0&1001&1001.0&521783.0&1001.0\\ 
608382&800&608019.0&1001&1001.0&608019.0&1001.0\\ 
693162&900&692765.0&921&920.0&692765.0&1001.0\\ 
777106&1000&776665.0&1001&1001.0&776665.0&1001.0\\ 
 \hline
\end{tabular}
\caption{Vanilla DQN}
\label{table:dqn_cartpole}
\end{table}
\begin{table}[h!]
\centering
\begin{tabular}{ |c|c|c|c|c|c|c| } 
 \hline
steps&episode&total return&episode len&episode return&raw return&best episode\\
\hline
122&1&121.0&122&121.0&121.0&121.0\\
151&2&149.0&29&28.0&149.0&121.0\\
248&3&245.0&97&96.0&245.0&121.0\\
331&4&327.0&83&82.0&327.0&121.0\\
359&5&354.0&28&27.0&354.0&121.0\\
388&6&382.0&29&28.0&382.0&121.0\\
418&7&411.0&30&29.0&411.0&121.0\\
460&8&452.0&42&41.0&452.0&121.0\\
551&9&542.0&91&90.0&542.0&121.0\\
636&10&626.0&85&84.0&626.0&121.0\\
872&12&860.0&111&110.0&860.0&124.0\\
1123&14&1109.0&111&110.0&1109.0&139.0\\
1432&17&1415.0&105&104.0&1415.0&139.0\\
1744&20&1724.0&114&113.0&1724.0&139.0\\
2481&25&2456.0&231&230.0&2456.0&230.0\\
3253&30&3223.0&163&162.0&3223.0&243.0\\
4610&40&4570.0&108&107.0&4570.0&300.0\\
6205&50&6155.0&86&85.0&6155.0&375.0\\
7805&60&7745.0&61&60.0&7745.0&375.0\\
9989&70&9920.0&86&85.0&9920.0&1001.0\\
15313&80&15238.0&238&237.0&15238.0&1001.0\\
19298&90&19214.0&80&79.0&19214.0&1001.0\\
21806&100&21713.0&28&27.0&21713.0&1001.0\\
27840&120&27729.0&216&215.0&27729.0&1001.0\\
34258&140&34129.0&383&382.0&34129.0&1001.0\\
46921&170&46769.0&180&179.0&46769.0&1001.0\\
60260&200&60081.0&877&876.0&60081.0&1001.0\\
80310&250&80089.0&334&333.0&80089.0&1001.0\\
105639&300&105378.0&159&158.0&105378.0&1001.0\\ 
153152&400&152812.0&159&158.0&152812.0&1001.0\\
199978&500&199559.0&260&259.0&199559.0&1001.0\\
258258&600&257765.0&1001&1001.0&257765.0&1001.0\\
319524&700&318964.0&202&201.0&318964.0&1001.0\\
372762&800&372128.0&167&166.0&372128.0&1001.0\\
418762&900&418047.0&377&376.0&418047.0&1001.0\\
475262&1000&474476.0&1001&1001.0&474476.0&1001.0\\
 \hline
\end{tabular}
\caption{Boot DQN}
\label{table:boot_dqn_cartpole}
\end{table}
\begin{table}[h!]
\centering
\begin{tabular}{ |c|c|c|c|c|c|c| } 
 \hline
steps&episode&total return&episode len&episode return&raw return&best episode\\
\hline
84&1&83.0&84&83.0&83.0&83.0\\
142&2&140.0&58&57.0&140.0&83.0\\
203&3&200.0&61&60.0&200.0&83.0\\
257&4&253.0&54&53.0&253.0&83.0\\
300&5&295.0&43&42.0&295.0&83.0\\
361&6&355.0&61&60.0&355.0&83.0\\
450&7&443.0&89&88.0&443.0&88.0\\
567&8&559.0&117&116.0&559.0&116.0\\
662&9&653.0&95&94.0&653.0&116.0\\
800&10&790.0&138&137.0&790.0&137.0\\
1030&12&1018.0&120&119.0&1018.0&137.0\\
1254&14&1240.0&131&130.0&1240.0&137.0\\
1654&17&1637.0&188&187.0&1637.0&187.0\\
1972&20&1952.0&135&134.0&1952.0&187.0\\
2578&25&2553.0&125&124.0&2553.0&187.0\\
3139&30&3109.0&106&105.0&3109.0&187.0\\
4198&40&4158.0&111&110.0&4158.0&187.0\\
5250&50&5200.0&101&100.0&5200.0&187.0\\
6272&60&6212.0&92&91.0&6212.0&187.0\\
7310&70&7240.0&40&39.0&7240.0&204.0\\
7811&80&7731.0&82&81.0&7731.0&204.0\\
9163&90&9073.0&160&159.0&9073.0&250.0\\
11091&100&10991.0&191&190.0&10991.0&306.0\\
14707&120&14587.0&164&163.0&14587.0&306.0\\
18564&140&18424.0&146&145.0&18424.0&369.0\\
25655&170&25485.0&111&110.0&25485.0&902.0\\
31176&200&30976.0&178&177.0&30976.0&902.0\\
44395&250&44145.0&178&177.0&44145.0&939.0\\
53754&300&53455.0&178&177.0&53455.0&1001.0\\
71189&400&70790.0&173&172.0&70790.0&1001.0\\
88763&500&88264.0&207&206.0&88264.0&1001.0\\
108915&600&108318.0&56&55.0&108318.0&1001.0\\
126820&700&126123.0&241&240.0&126123.0&1001.0\\
143856&800&143059.0&94&93.0&143059.0&1001.0\\
157030&900&156133.0&156&155.0&156133.0&1001.0\\
173286&1000&172289.0&166&165.0&172289.0&1001.0\\
 \hline
\end{tabular}
\caption{A2c Cartpole}
\label{table:a2c_cartpole}
\end{table}
\begin{table}[h!]
\centering
\begin{tabular}{ |c|c|c|c|c|c| } 
 \hline
steps&episode&total return&episode len&episode return&raw return\\
\hline
1001&1&-1001.0&1001&-1001.0&-1001.0\\
2002&2&-2002.0&1001&-1001.0&-2002.0\\
3003&3&-3003.0&1001&-1001.0&-3003.0\\
3100&4&-3100.0&97&-97.0&-3100.0\\
4101&5&-4101.0&1001&-1001.0&-4101.0\\
5102&6&-5102.0&1001&-1001.0&-5102.0\\
5158&7&-5158.0&56&-56.0&-5158.0\\
5236&8&-5236.0&78&-78.0&-5236.0\\
5287&9&-5287.0&51&-51.0&-5287.0\\
5353&10&-5353.0&66&-66.0&-5353.0\\
6384&12&-6384.0&1001&-1001.0&-6384.0\\
7421&14&-7421.0&1001&-1001.0&-7421.0\\
7582&17&-7582.0&48&-48.0&-7582.0\\
8708&20&-8708.0&47&-47.0&-8708.0\\
9867&25&-9867.0&38&-38.0&-9867.0\\
12050&30&-12050.0&71&-71.0&-12050.0\\
14427&40&-14427.0&29&-29.0&-14427.0\\
16779&50&-16779.0&52&-52.0&-16779.0\\
17204&60&-17204.0&44&-44.0&-17204.0\\
17569&70&-17569.0&28&-28.0&-17569.0\\
17951&80&-17951.0&44&-44.0&-17951.0\\
18289&90&-18289.0&41&-41.0&-18289.0\\
18639&100&-18639.0&39&-39.0&-18639.0\\
19311&120&-19311.0&44&-44.0&-19311.0\\
19999&140&-19999.0&32&-32.0&-19999.0\\
21105&170&-21105.0&37&-37.0&-21105.0\\
22137&200&-22137.0&27&-27.0&-22137.0\\
23863&250&-23863.0&47&-47.0&-23863.0\\
25514&300&-25514.0&30&-30.0&-25514.0\\
29048&400&-29048.0&29&-29.0&-29048.0\\
32763&500&-32763.0&51&-51.0&-32763.0\\
36452&600&-36452.0&35&-35.0&-36452.0\\
40589&700&-40589.0&40&-40.0&-40589.0\\
43959&800&-43959.0&28&-28.0&-43959.0\\
47403&900&-47403.0&37&-37.0&-47403.0\\
50901&1000&-50901.0&37&-37.0&-50901.0\\
 \hline
\end{tabular}
\caption{DQN Moutnain Car}
\label{table:dqn_mountain_car}
\end{table}
\begin{table}[h!]
\centering
\begin{tabular}{ |c|c|c|c|c|c| } 
 \hline
steps&episode&total return&episode len&episode return&raw return\\
\hline
1001&1&-1001.0&1001&-1001.0&-1001.0\\
2002&2&-2002.0&1001&-1001.0&-2002.0\\
2040&3&-2040.0&38&-38.0&-2040.0\\
2076&4&-2076.0&36&-36.0&-2076.0\\
3077&5&-3077.0&1001&-1001.0&-3077.0\\
3115&6&-3115.0&38&-38.0&-3115.0\\
3152&7&-3152.0&37&-37.0&-3152.0\\
3193&8&-3193.0&41&-41.0&-3193.0\\
3230&9&-3230.0&37&-37.0&-3230.0\\
3258&10&-3258.0&28&-28.0&-3258.0\\
3325&12&-3325.0&30&-30.0&-3325.0\\
3384&14&-3384.0&30&-30.0&-3384.0\\
4470&17&-4470.0&47&-47.0&-4470.0\\
4595&20&-4595.0&35&-35.0&-4595.0\\
4783&25&-4783.0&28&-28.0&-4783.0\\
4997&30&-4997.0&44&-44.0&-4997.0\\
5513&40&-5513.0&50&-50.0&-5513.0\\
5934&50&-5934.0&30&-30.0&-5934.0\\
6326&60&-6326.0&44&-44.0&-6326.0\\
6678&70&-6678.0&28&-28.0&-6678.0\\
7057&80&-7057.0&48&-48.0&-7057.0\\
7439&90&-7439.0&48&-48.0&-7439.0\\
7840&100&-7840.0&47&-47.0&-7840.0\\
8719&120&-8719.0&40&-40.0&-8719.0\\
9488&140&-9488.0&42&-42.0&-9488.0\\
10522&170&-10522.0&38&-38.0&-10522.0\\
11570&200&-11570.0&31&-31.0&-11570.0\\
13289&250&-13289.0&31&-31.0&-13289.0\\
15031&300&-15031.0&28&-28.0&-15031.0\\
18448&400&-18448.0&35&-35.0&-18448.0\\
21934&500&-21934.0&38&-38.0&-21934.0\\
25491&600&-25491.0&29&-29.0&-25491.0\\
29021&700&-29021.0&33&-33.0&-29021.0\\
32592&800&-32592.0&31&-31.0&-32592.0\\
36315&900&-36315.0&35&-35.0&-36315.0\\
39884&1000&-39884.0&42&-42.0&-39884.0\\
 \hline
\end{tabular}
\caption{Boot Moutnain Car}
\label{table:boot_mountain_car}
\end{table}
\begin{table}[h!]
\centering
\begin{tabular}{ |c|c|c|c|c|c| } 
 \hline
steps&episode&total return&episode len&episode return&raw return\\
\hline
83&1&-83.0&83&-83.0&-83.0\\
121&2&-121.0&38&-38.0&-121.0\\
184&3&-184.0&63&-63.0&-184.0\\
216&4&-216.0&32&-32.0&-216.0\\
255&5&-255.0&39&-39.0&-255.0\\
292&6&-292.0&37&-37.0&-292.0\\
322&7&-322.0&30&-30.0&-322.0\\
365&8&-365.0&43&-43.0&-365.0\\
416&9&-416.0&51&-51.0&-416.0\\
448&10&-448.0&32&-32.0&-448.0\\
507&12&-507.0&30&-30.0&-507.0\\
591&14&-591.0&40&-40.0&-591.0\\
685&17&-685.0&28&-28.0&-685.0\\
783&20&-783.0&30&-30.0&-783.0\\
941&25&-941.0&27&-27.0&-941.0\\
1107&30&-1107.0&36&-36.0&-1107.0\\
1447&40&-1447.0&44&-44.0&-1447.0\\
1817&50&-1817.0&35&-35.0&-1817.0\\
2143&60&-2143.0&31&-31.0&-2143.0\\
2455&70&-2455.0&31&-31.0&-2455.0\\
2806&80&-2806.0&33&-33.0&-2806.0\\
3151&90&-3151.0&35&-35.0&-3151.0\\
3498&100&-3498.0&31&-31.0&-3498.0\\
4165&120&-4165.0&44&-44.0&-4165.0\\
4831&140&-4831.0&32&-32.0&-4831.0\\
5784&170&-5784.0&30&-30.0&-5784.0\\
6775&200&-6775.0&41&-41.0&-6775.0\\
8493&250&-8493.0&31&-31.0&-8493.0\\
10151&300&-10151.0&43&-43.0&-10151.0\\
13532&400&-13532.0&29&-29.0&-13532.0\\
16932&500&-16932.0&35&-35.0&-16932.0\\
20328&600&-20328.0&35&-35.0&-20328.0\\
23642&700&-23642.0&35&-35.0&-23642.0\\
27004&800&-27004.0&36&-36.0&-27004.0\\
30343&900&-30343.0&30&-30.0&-30343.0\\
33735&1000&-33735.0&30&-30.0&-33735.0\\
 \hline
\end{tabular}
\caption{A2C Moutnain Car}
\label{table:a2c_mountain_car}
\end{table}
\begin{table}[h!]
\centering
\begin{tabular}{ |c|c|c|c|c|c|c| } 
 \hline
steps&ep.e&total return&ep. len&episode return&ttl. bad ep.&denoised ret\\
\hline
18&1&-0.005&18&-0.005&1&0\\
36&2&-0.0116666666666&18&-0.0066666666667&2&0\\
54&3&-0.0183333333333&18&-0.0066666666667&3&0\\
72&4&-0.023888888889&18&-0.0055555555556&4&0\\
90&5&-0.0294444444447&18&-0.0055555555556&5&0\\
108&6&-0.036111111109&18&-0.0066666666667&6&0\\
126&7&-0.041666666661&18&-0.0055555555556&7&0\\
144&8&-0.046666666658&18&-0.005&8&0\\
162&9&-0.0511111111996&18&-0.00444444444444&9&0\\
180&10&-0.059444444428&18&-0.0083333333333&10&0\\
216&12&-0.069444444431&18&-0.005&12&0\\
252&14&-0.080555555549&18&-0.0066666666667&14&0\\
306&17&-0.093333333335&18&-0.005&17&0\\
360&20&-0.107777777789&18&-0.0055555555556&20&0\\
450&25&-0.138333333364&18&-0.0055555555556&25&0\\
540&30&-0.164444444492&18&-0.005&30&0\\
720&40&-0.207222222297&18&-0.00333333333335&40&0\\
900&50&-0.24888888899&18&-0.00444444444444&50&0\\
1080&60&-0.28166666663&18&-0.00222222222222&60&0\\
1260&70&-0.31888888869&18&-0.00333333333335&70&0\\
1440&80&-0.35444444409&18&-0.00333333333335&80&0\\
1620&90&-0.381666666194&18&-0.00222222222222&90&0\\
1800&100&-0.419444443807&18&-0.00444444444444&100&0\\
2160&120&-0.45722222142&18&-0.00111111111111&120&0\\
2520&140&-0.50555555554&18&-0.00166666666666&140&0\\
3060&170&-0.57611111179&18&-0.00111111111111&170&0\\
3600&200&-0.62166666615&18&-0.00166666666666&200&0\\
4500&250&-0.69666666482&18&-0.00111111111111&250&0\\
5400&300&-0.75055555348&18&-0.00055555555556&300&0\\
7200&400&-0.87111111851&18&-0.00166666666666&400&0\\
9000&500&-0.96833333031&18&-0.00055555555556&500&0\\
10800&600&-1.08611110928&18&-0.00222222222222&600&0\\
12600&700&-1.174444444&18&0.0&700&0\\
14400&800&-1.31666666844&18&-0.00111111111111&800&0\\
16200&900&-1.39444444744&18&-0.0027777777778&900&0\\
18000&1000&-1.4772222265&18&-0.0027777777778&1000&0\\
21600&1200&-1.91777778895&18&-0.00222222222222&1200&0\\
25200&1400&-2.19055557097&18&0.0&1400&0\\
30600&1700&-2.7011111345&18&-0.00111111111111&1700&0\\
36000&2000&-3.40111114545&18&-0.00111111111111&2000&0\\
45000&2500&-4.4338888894&18&-0.0027777777778&2500&0\\
54000&3000&-5.4888888859&18&-0.00222222222222&3000&0\\
72000&4000&-7.5700000095&18&-0.00166666666666&4000&0\\
90000&5000&-9.6744444468&18&-0.0027777777778&5000&0\\
108000&6000&-11.7505557204&18&-0.00444444444444&6000&0\\
126000&7000&-13.886666865&18&-0.0027777777778&7000&0\\
144000&8000&-16.165000081&18&-0.00111111111111&8000&0\\
162000&9000&-18.4572227062&18&-0.00222222222222&9000&0\\
180000&10000&-20.8227775424&18&-0.00111111111111&10000&0\\
 \hline
\end{tabular}
\caption{DQN Deep Sea}
\label{table:dqn_deep_sea}
\end{table}
\begin{table}[h!]
\centering
\begin{tabular}{ |c|c|c|c|c|c|c| } 
 \hline
steps&ep.e&total return&ep. len&episode return&ttl. bad ep.&denoised ret\\
\hline
10&1&-0.005&10&-0.005&1&0\\
20&2&-0.008&10&-0.003&2&0\\
30&3&-0.013000000000000005&10&-0.005&3&0\\
40&4&-0.01800000000000001&10&-0.005&4&0\\
50&5&-0.023000000000000013&10&-0.005&5&0\\
60&6&-0.02900000000000002&10&-0.006&6&0\\
70&7&-0.03200000000000002&10&-0.003&7&0\\
80&8&-0.037000000000000026&10&-0.005&8&0\\
90&9&-0.04200000000000003&10&-0.005&9&0\\
100&10&-0.04900000000000004&10&-0.007&10&0\\
120&12&-0.06200000000000005&10&-0.008&12&0\\
140&14&-0.07300000000000005&10&-0.005&14&0\\
170&17&-0.08600000000000006&10&-0.004&17&0\\
200&20&-0.10000000000000007&10&-0.006&20&0\\
250&25&-0.12800000000000009&10&-0.005&25&0\\
300&30&-0.1500000000000001&10&-0.004&30&0\\
400&40&-0.19100000000000014&10&-0.006&40&0\\
500&50&-0.22900000000000018&10&-0.005&50&0\\
600&60&-0.2840000000000002&10&-0.008&60&0\\
700&70&-0.32900000000000024&10&-0.005&70&0\\
800&80&-0.3720000000000003&10&-0.002&80&0\\
900&90&-0.4250000000000003&10&-0.005&90&0\\
1000&100&-0.4870000000000004&10&-0.008&100&0\\
1200&120&-0.5780000000000004&10&-0.004&120&0\\
1400&140&-0.6680000000000005&10&-0.003&140&0\\
1700&170&-0.7930000000000006&10&-0.003&170&0\\
2000&200&-0.9120000000000007&10&-0.004&200&0\\
2500&250&-1.078999999999992&10&-0.003&250&0\\
3000&300&-1.1729999999999816&10&-0.001&300&0\\
4000&400&-1.3039999999999672&10&0.0&400&0\\
5000&500&-1.4129999999999552&10&-0.001&500&0\\
6000&600&-1.513999999999944&10&-0.001&600&0\\
7000&700&-1.6159999999999328&10&-0.001&700&0\\
8000&800&-1.7219999999999211&10&-0.001&800&0\\
9000&900&-1.82399999999991&10&-0.001&900&0\\
10000&1000&-1.923999999999899&10&-0.001&1000&0\\
12000&1200&-2.1219999999998773&10&-0.001&1200&0\\
14000&1400&-2.1819999999998707&10&0.0&1400&0\\
17000&1700&-2.18999999999987&10&0.0&1700&0\\
20000&2000&-2.18999999999987&10&0.0&2000&0\\
25000&2500&-2.2019999999998685&10&0.0&2500&0\\
30000&3000&-2.232999999999865&10&0.0&3000&0\\
40000&4000&-2.2379999999998645&10&0.0&4000&0\\
50000&5000&-3.132999999999766&10&-0.002&5000&0\\
60000&6000&-4.163999999999725&10&-0.001&6000&0\\
70000&7000&-5.16600000000006&10&-0.001&7000&0\\
80000&8000&-6.1710000000003955&10&-0.001&8000&0\\
90000&9000&-7.175000000000731&10&-0.001&9000&0\\
100000&10000&-9.083000000000405&10&-0.003&10000&0\\
 \hline
\end{tabular}
\caption{A2C Deep Sea}
\label{table:a2c_deep_sea}
\end{table}
\begin{table}[h!]
\centering
\begin{tabular}{ |c|c|c|c|c|c|c| } 
 \hline
steps&ep.e&total return&ep. len&episode return&ttl. bad ep.&denoised ret\\
\hline
10&1&-0.005&10&-0.005&1&0.0\\
20&2&-0.011000000000000003&10&-0.006&2&0.0\\
30&3&-0.014000000000000005&10&-0.003&3&0.0\\
40&4&-0.02000000000000001&10&-0.006&4&0.0\\
50&5&-0.025000000000000015&10&-0.005&5&0.0\\
60&6&-0.03000000000000002&10&-0.005&6&0.0\\
70&7&-0.036000000000000025&10&-0.006&7&0.0\\
80&8&-0.04100000000000003&10&-0.005&8&0.0\\
90&9&-0.04500000000000003&10&-0.004&9&0.0\\
100&10&-0.05000000000000004&10&-0.005&10&0.0\\
120&12&-0.059000000000000045&10&-0.004&12&0.0\\
140&14&-0.06900000000000005&10&-0.005&14&0.0\\
170&17&-0.08300000000000006&10&-0.004&17&0.0\\
200&20&-0.09800000000000007&10&-0.005&20&0.0\\
250&25&-0.11900000000000009&10&-0.004&25&0.0\\
300&30&-0.1410000000000001&10&-0.007&30&0.0\\
400&40&-0.19300000000000014&10&-0.004&40&0.0\\
500&50&-0.2460000000000002&10&-0.007&50&0.0\\
600&60&-0.3080000000000002&10&-0.007&60&0.0\\
700&70&-0.36800000000000027&10&-0.007&70&0.0\\
800&80&-0.43400000000000033&10&-0.00900001&80&0.0\\
900&90&-0.5070000000000003&10&-0.008&90&0.0\\
1000&100&-0.5750000000000004&10&-0.00900001&100&0.0\\
1200&120&-0.7150000000000005&10&-0.005&120&0.0\\
1400&140&-0.8220000000000006&10&-0.005&140&0.0\\
1700&170&3.9960000000000027&10&0.99&165&5.0\\
2000&200&30.70999999999987&10&0.99&168&32.0\\
2500&250&80.20999999999982&10&0.99&168&82.0\\
3000&300&129.70999999999745&10&0.99&168&132.0\\
4000&400&228.70999999999268&10&0.99&168&232.0\\
5000&500&327.7100000000084&10&0.99&168&332.0\\
6000&600&426.71000000003204&10&0.99&168&432.0\\
7000&700&525.7100000000556&10&0.99&168&532.0\\
8000&800&624.7100000000793&10&0.99&168&632.0\\
9000&900&723.7100000001029&10&0.99&168&732.0\\
10000&1000&822.7100000001266&10&0.99&168&832.0\\
12000&1200&1020.7100000001739&10&0.99&168&1032.0\\
14000&1400&1218.710000000221&10&0.99&168&1232.0\\
17000&1700&1515.710000000292&10&0.99&168&1532.0\\
20000&2000&1811.711000000363&10&0.99&169&1831.0\\
25000&2500&2306.7109999998875&10&0.99&169&2331.0\\
30000&3000&2801.710999998869&10&0.99&169&2831.0\\
40000&4000&3791.7109999968316&10&0.99&169&3831.0\\
50000&5000&4781.710999994794&10&0.99&169&4831.0\\
60000&6000&5771.710999992757&10&0.99&169&5831.0\\
70000&7000&6761.71099999072&10&0.99&169&6831.0\\
80000&8000&7751.7109999886825&10&0.99&169&7831.0\\
90000&9000&8741.710999986646&10&0.99&169&8831.0\\
100000&10000&9731.710999984609&10&0.99&169&9831.0\\
 \hline
\end{tabular}
\caption{BootDQN Deep Sea}
\label{table:boot_deep_sea}
\end{table} 

\addcontentsline{toc}{chapter}{Bibliography}

\bibliographystyle{apalike}
\bibliography{biblio}

\end{document}